\documentclass[nohyperref]{article}

\usepackage[usenames, dvipsnames]{color}
\usepackage[dvipsnames, svgnames, x11names]{xcolor}

\usepackage{microtype}
\usepackage{graphicx}
\usepackage{subfigure}
\usepackage{booktabs} % for professional tables

\usepackage{hyperref}
\usepackage{enumitem}

\usepackage[accepted]{icml2022}

\usepackage{amsmath}
\usepackage{amssymb}
\usepackage{mathtools}
\usepackage{amsthm}

\usepackage[capitalize,noabbrev]{cleveref}

\theoremstyle{plain}

\theoremstyle{definition}

\theoremstyle{remark}

\usepackage[textsize=tiny]{todonotes}

\icmltitlerunning{Learning Domain Adaptive Object Detection with Probabilistic Teacher}

\usepackage{threeparttable}
\usepackage{multirow}
\usepackage{listings}
\usepackage{colortbl}
\usepackage{bbding}
\usepackage{wrapfig}

\usepackage{newfloat}
\floatstyle{ruled}
\newfloat{listing}{tb}{lst}{}
\floatname{listing}{Listing}

\usepackage{eqparbox}
\renewcommand{\algorithmiccomment}[1]{\hfill\eqparbox{COMMENT}{\# #1}}
\begin{document}

\twocolumn[
\icmltitle{Learning Domain Adaptive Object Detection with Probabilistic Teacher}



\icmlsetsymbol{intern}{*}
\icmlsetsymbol{mentor}{*}

\begin{icmlauthorlist}
\icmlauthor{Meilin Chen}{intern,zju}
\icmlauthor{Weijie Chen}{mentor,zju,hk}
\icmlauthor{Shicai Yang}{hk}
\icmlauthor{Jie Song}{zju}
\icmlauthor{Xinchao Wang}{nus}
\icmlauthor{Lei Zhang}{cqu} \\
\icmlauthor{Yunfeng Yan}{zju}
\icmlauthor{Donglian Qi}{hn,zju}
\icmlauthor{Yueting Zhuang}{zju}
\icmlauthor{Di Xie}{hk}
\icmlauthor{Shiliang Pu}{hk}\\
Code available at \url{https://github.com/hikvision-research/ProbabilisticTeacher}
\end{icmlauthorlist}

\icmlaffiliation{zju}{Zhejiang University}
\icmlaffiliation{hn}{Hainan Institute of Zhejiang University}
\icmlaffiliation{hk}{Hikvision Research Institute}
\icmlaffiliation{nus}{National University of Singapore}
\icmlaffiliation{cqu}{Chongqing University}
\icmlcorrespondingauthor{Yueting Zhuang}{yzhuang@zju.edu.cn}
\icmlcorrespondingauthor{Shiliang Pu}{pushiliang.hri@hikvision.com}

\icmlkeywords{Domain Adaptive Object Detection, Self-Training}

\vskip 0.3in
]
\printAffiliationsAndNotice{* This work was done when Meilin Chen (merlinis@zju.edu.cn) was an intern in Hikvision Research Institute supervised by Weijie Chen (chenweijie5@hikvision.com). }

\begin{abstract}
Self-training for unsupervised domain adaptive object detection is a challenging task, 
of which the performance depends heavily on the quality of pseudo boxes. Despite the promising results, prior works have largely overlooked the uncertainty of pseudo boxes during self-training. In this paper, we present a simple yet effective framework, 
termed as \emph{\textbf{P}robabilistic \textbf{T}eacher} (\textbf{PT}), which aims to capture the uncertainty of unlabeled target data from a gradually evolving teacher and 
guides the learning of a student in a mutually beneficial manner. 
Specifically, we propose to leverage the uncertainty-guided consistency training to promote classification adaptation and localization adaptation, rather than filtering pseudo boxes via an elaborate confidence threshold. In addition, we conduct anchor adaptation in parallel with localization adaptation, since anchor can be regarded as a learnable parameter. Together with this framework, we also present a novel \emph{\textbf{E}ntropy \textbf{F}ocal \textbf{L}oss}~(\textbf{EFL}) to further facilitate the uncertainty-guided self-training. 
Equipped with EFL, PT outperforms all previous baselines by a large margin and achieve new state-of-the-arts.
\end{abstract}
\vspace{-1em}

\section{Introduction}

Convolutional neural networks have shown remarkable performance
for object detection when trained on large-scale and high-quality annotated data. 
However, when deployed to 
unseen data, the detector dramatically degrades due to
domain shifts such as weather changes, light conditions variations, or image corruptions \cite{2}. 
To remedy this issue, 
unsupervised domain adaptive object detection (UDA-OD) methods
have been proposed~\cite{3,4,6,7,8,9,10,11,SFOD},
whose goal 
is to transfer pre-trained models from a labeled source domain to an unlabeled target domain with different data distribution. 
Recently, UDA-OD methods have witnessed a strong demands
in real-world scenarios such as automatic driving and edge AI, 
where domain shifts are common and collecting high-quality annotated target data is expensive.

\begin{figure}[tbp]
    \centering
    \includegraphics[width=0.46\textwidth]{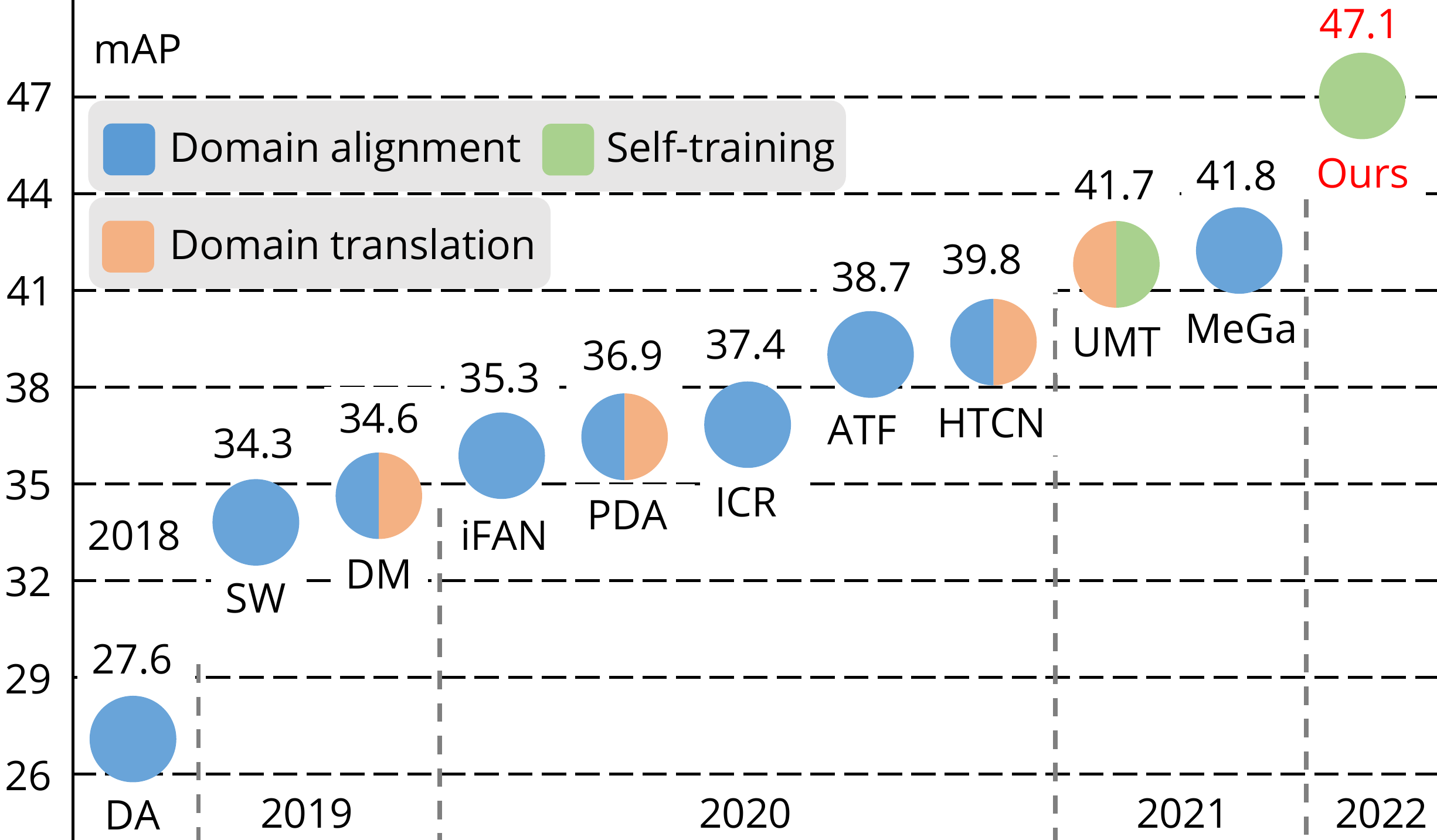}
    \vskip -0.1in
    \caption{Performance comparison of adapting different models from normal to foggy weather. Our framework achieves the state-of-the-art result by simply adopting a self-training mechanism. %\wj{2022?}
    }
    \label{method_comp}
    \vskip -0.2in
\end{figure}

As shown in Fig.\ref{method_comp}, various methods have been proposed for this task, and they can be categorized as domain alignment, domain translation and self-training methods. Domain alignment aims to learn domain-invariant representation using domain classifiers and gradient reversal layers \cite{3,4,6}.  Domain translation, on the other hand, attempts to translate the labeled source data into target-like styles to drive adaptation training \cite{7,8,9}. Very recently, self-training is proposed to leverage teacher-student mutual learning to progressively improve the performance of unlabeled target data \cite{10,11}. 
Specifically, self-training removes the necessity of 
extra training paradigms like adversarial training and style transfer,
and has recently  demonstrated promising results.
Our proposed method, as will be detailed in the later sections, 
falls into the self-training  category. 

\begin{figure}[tbp]
    \centering
    \includegraphics[width=0.45\textwidth]{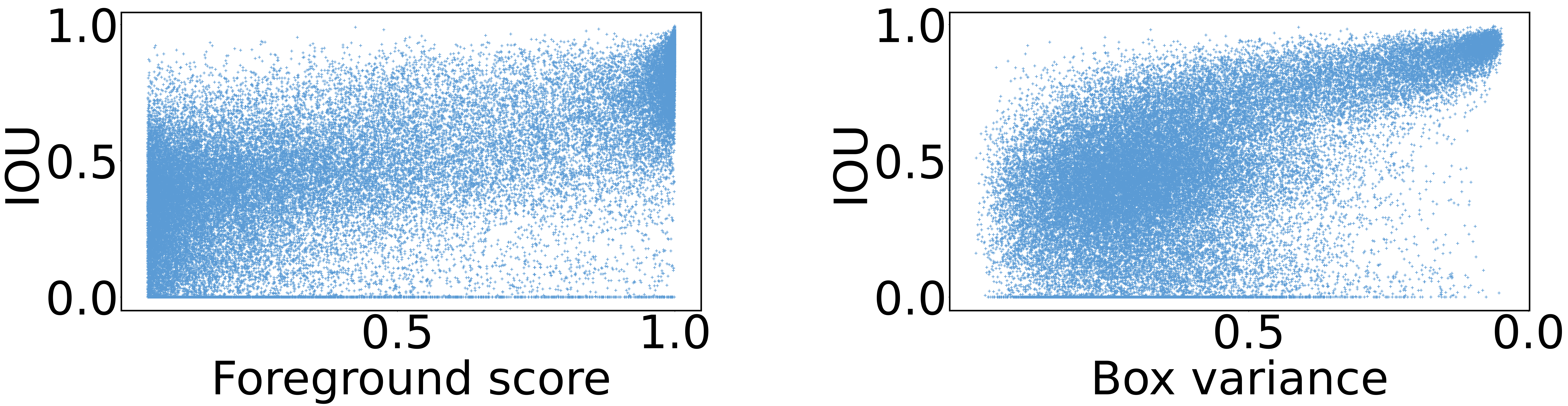}
\vspace{-1em}
    \caption{Box variance $\sigma^{2}$ is better than foreground score to measure the localization accuracy of bboxes (IoU with ground-truth bboxes). Results on the Foggy Cityscapes are presented.}
    \label{sigma_score_iou}
\vspace{-2em}
\end{figure}

The critical component in self-training lies in the pseudo labeling. A popular solution is to filter pseudo boxes via an elaborate category confidence threshold \cite{10,11}. However, there are two inherent challenges in this paradigm. \textit{Dependence challenge}: the performance in this paradigm depends heavily on the selection of the threshold, while in many if not all cases, no annotated target data is available to tune the confidence threshold. \textit{Performance challenge}: since only category confidence is considered while localization confidence is not, this simple solution cannot guarantee the quality of pseudo boxes (see Fig.\ref{sigma_score_iou}). 

To address these issues, from the perspective of uncertainty, we present in this paper a threshold-free framework, termed as \emph{\textbf{P}robabilistic \textbf{T}eacher} (\textbf{PT}), to apply cross-domain self-training via uncertainty-guided consistency training between the teacher and student models for both classification and localization adaptation. In our proposed framework, the existing Faster-RCNN \cite{23} is restructured into a probabilistic one, since the existing Faster-RCNN is incapable to predict localization uncertainty. 
In this way, both category and localization labels can be represented as probability distributions,  providing a ground for the teacher model to annotate target pseudo boxes with uncertainty.

Furthermore, another issue in prior work comes to the anchor. As a scene-sensitive parameter, anchor shapes have to be manually tweaked to improve accuracy when applying anchor-based detector to a specific object detection dataset. In the existing works, source and target domains usually share the same anchors. However, source and target domains generally have different distributions of bounding box (bbox) sizes due to domain shifts. In this paper, we propose to carry out anchor adaptation in parallel with localization adaptation. Thus, our approach unifies classification, localization and anchor adaptations into one framework.

To further facilitate uncertainty-guided self-training, we design an \emph{\textbf{E}ntropy \textbf{F}ocal \textbf{L}oss} (\textbf{EFL}) to drive uncertainty-guided consistency training in both classification and localization branches, encouraging the model to pay more attention to the lower-entropy pseudo boxes.

Compared with the existing self-training methods, our approach does not require filtering the target pseudo boxes using a carefully fine-tuned confidence threshold.
This makes our model particularly suitable for the UDA-OD setting, where no annotated target data is available for the filtering threshold tuning.
Of particular importance, our PT approach can be seamlessly and effortlessly extended to source-free UDA-OD setting (privacy-critical scenario), where only unlabeled target data is involved into self-training for the purpose of privacy protection, achieving remarkable improvements compared with previous approaches.

What is worth highlighting that we draw several interesting yet novel findings via extensive ablation studies: 1) Strong data augmentation is an implicit intra-domain alignment method to bridge the intra-domain gap between the true labels and false labels in target domain. 2) Data augmentation plays a much more important role in self-training approaches than domain alignment counterparts. 3) Merely adopting localization adaptation alone can still improve the adaptation performance against ``source only'' remarkably.

We summarize the takeaways as well as contributions as: 
\begin{itemize}[leftmargin=12pt, topsep=2pt, itemsep=0pt]
\item We propose a \textit{threshold-free} framework to explore cross-domain object detection via an uncertainty-driven self-training paradigm. It firstly unifies classification, localization as well as anchor adaptations into one framework.
\item We design an EFL loss for PT framework to further facilitate uncertainty-guided cross-domain self-training.
\item We draw several interesting yet novel experimental findings, which can inspire the future works in UDA-OD.
\item Our framework achieves the new state-of-the-art results on multiple source-based / free UDA-OD benchmarks, and surpasses previous approaches by a large margin. 
\end{itemize}

\begin{figure*}[ht]
    \centering
    \includegraphics[width=0.97\textwidth]{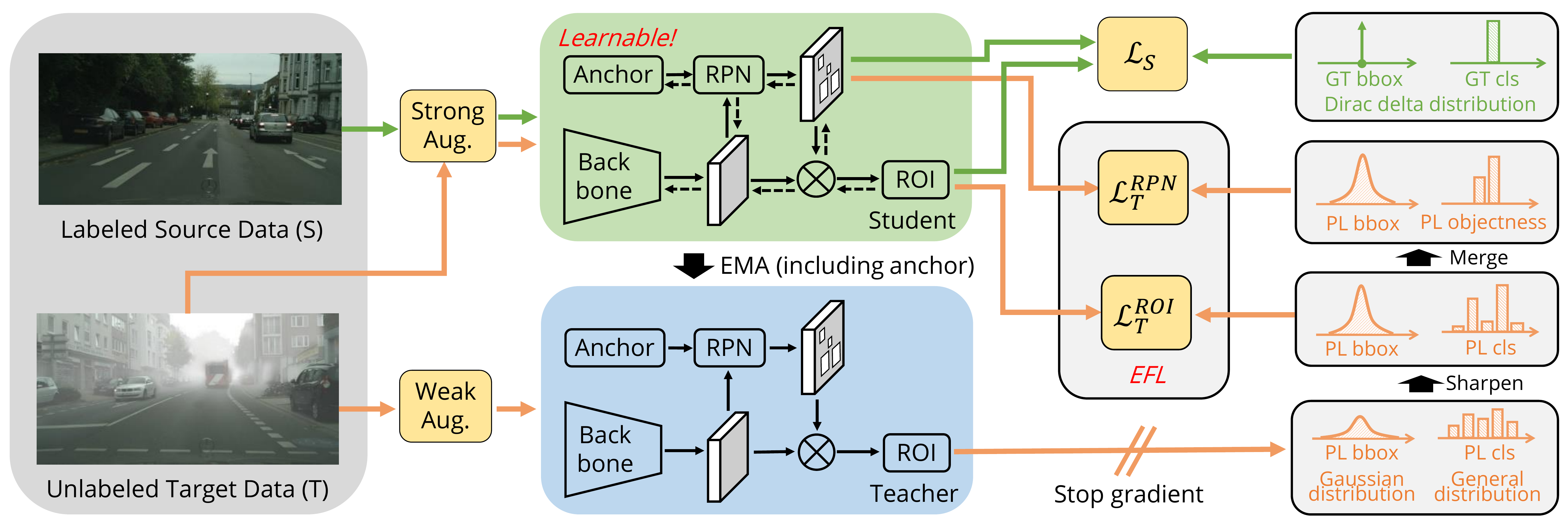}
    \vspace{-1em}
    \caption{Illustration of the proposed Probabilistic Teacher. Unlabeled target data with weak augmentation are fed into the teacher model to generate pseudo boxes, which contain classification and localization probability distributions. Both labeled source data and pseudo-labeled target data with sequent sharpening operation are used to train the student model via uncertainty-guided consistency training with strong data augmentation. Also, anchor adaptation is conducted in parallel with localization adaptation. To promote PT framework, Entropy Focal Loss (EFL) is proposed to further facilitate the cross-domain self-training.}
    \label{fig_pipline}
\vskip -0.1in
\end{figure*}

\section{Related Works}
\noindent\textbf{Unsupervised Domain Adaptive Object Detection}
Several approaches have been proposed for UDA-OD, which can be categorized into domain alignment, domain translation and self-training methods. These methods have been introduced briefly in the section above. We discuss the last one more detailedly since our work is built on self-training. As mentioned above, the most critical component in self-training is to exploit pseudo boxes. To obtain more accurate pseudo boxes, Unbiased Mean Teacher \cite{10} translates the target domain into a source-like one to generate pseudo boxes. In contrast, SimROD \cite{11} introduces a teacher model with a larger capacity to generate pseudo boxes. However, these existing works try to generate more accurate pseudo boxes using different techniques while the pseudo boxes are inevitable to be noisy. To address this problem, we propose an uncertainty-guided framework to deal with noisy pseudo boxes dynamically during cross-domain self-training.

\noindent\textbf{Self Training for Object Detection}
Self-training methods have been explored in many previous works for semi-supervised object detection~\cite{12,13,14,15,LabelMatch2022,weihaoCVPR22,yidingCVPR20,yidingNeurIPS20,suchengCVPR22}, in which pseudo boxes of unlabeled data are filtered to train the detectors using a carefully fine-tuned category confidence threshold. STAC \cite{13} is proposed to pre-train a detector using a small amount of labeled data and then generate pseudo boxes on unlabeled data to fine-tune the pre-trained detector. However, the pseudo boxes are generated only once and fixed throughout the rest of the training progress. Unbiased Teacher \cite{14} attempts to update the pseudo boxes via a mean teacher mechanism, while the box regression is only performed on the labeled data. Soft Teacher \cite{15} is proposed to use a box jitter method to measure the localization accuracy to filter the pseudo boxes. LabelMatch \cite{LabelMatch2022} exploits the Label Distribution Consistency assumption between labeled and unlabeled data to update the confidence threshold for pseudo labeling. It is reasonable to search for an appropriate confidence threshold using an annotated validation set for semi-supervised detection. However, no annotated target data is available for the filtering threshold tuning during UDA-OD, which inspires us to exploit the uncertainty of pseudo boxes rather than filtering the pseudo boxes via an elaborate confidence threshold.

\section{Preliminary}
Faster-RCNN, a benchmark detector for UDA-OD task, decouples object detection into a cross-entropy-based classification branch and a $\mathcal{L}$1-based bbox regression branch. In the classification branch, the predicted probability distribution over label space is natural to capture the classification uncertainty, while the Dirac delta modeling for bbox regression makes it incapable to obtain the localization uncertainty. To remedy this, in this section, we augment the existing Faster-RCNN detector into a probabilistic one, dubbed as Probabilistic Faster-RCNN, where both category and localization labels are represented as probability distributions.

Concretely, each coordinate ($t_x$, $t_y$, $t_w$, and $t_h$) of a bbox can be modeled as a single Gaussian model \cite{16}. Let coordinate $t$ be an univariate Gaussian distributed random variable parameterized by mean $\mu$ and variance $ \sigma^{2}$: $ t \sim \mathcal{N}\left(\mu, \sigma^{2}\right)$. Note that $\sigma^{2}$ is constrained as a value between zero and one with a sigmoid function. In this way, bbox regression loss can be implemented by a cross-entropy function between the ground-truth distribution $t^{G T}$ (Dirac delta one) and the predicted one $t$ (Gaussian one):
\begin{equation}
\begin{small}
\begin{aligned}
\mathcal{L}_{bbox}&=\frac{1}{N_{bbox}} \sum_{i} \mathbb{I}_{fg}(t_{i}) \mathcal{H}(t_{i}^{G T}, t_{i}) \\
& \overset{\bigstar}{=}-\frac{1}{N_{bbox}} \sum_{i} \mathbb{I}_{fg}(t_{i})\log (\mathcal{N}(t_{i}^{G T} ; \mu_{i},\sigma_{i}^{2}))
\label{equation3}
\end{aligned}
\end{small}
\end{equation}
where $\mathcal{H}(\cdot, \cdot)$ denotes the standard cross-entropy function. $t_{i}^{G T}$ is the ground-truth bbox coordinate associated with the  $i_{th}$  predicted bbox $t_{i}$. $N_{bbox}$ is the number of anchors or proposals. $\mathbb{I}_{fg}(\cdot)$ is a sign function to indicate whether the predicted bbox is matched to an anchor or proposal. $\mu_{i}$ and ${\sigma_{i}}^{2}$ are the predicted coordinate mean and variance. $\mathcal{N}(t_{i}^{G T} ; \mu_{i},\sigma_{i}^{2})$ denotes the probability of $t_{i}^{G T}$ in the Gaussian distribution. 
As shown in Fig.\ref{sigma_score_iou}, ${\sigma_{i}}^{2}$ is better than foreground score to measure the localization accuracy. The detailed proof of step $\bigstar$ can be found in the appendix \ref{appendix_Dirac_Gaussian}.

With the probabilistic modeling for bbox regression, Probabilistic Faster-RCNN is able to capture the uncertainty of both classification and localization for each prediction. The overall training objective can be reformulated as:
\begin{equation}
\begin{small}
\begin{aligned}
\mathcal{L}_{S}=\mathcal{L}_{cls}^{RPN}+\mathcal{L}_{cls}^{ROI}+\mathcal{L}_{bbox}^{RPN}+\mathcal{L}_{bbox}^{ROI}
\label{source-loss}
\end{aligned}
\end{small}
\end{equation}
where all the four terms are cross-entropy losses, and equally weighted following the original Faster-RCNN. With the favor of Probabilistic Faster-RCNN, the proposed method, Probabilistic Teacher, is presented in the next section.

\section{Probabilistic Teacher}

\subsection{Overview}
We depict the overview of Probabilistic Teacher in Fig.\ref{fig_pipline}. Probabilistic Teacher contains two training steps, Pretraining and Mutual learning. \textbf{1) Pretraining}. We train the detector using the labeled source data to initialize the detector, and then duplicate the trained weights to both the teacher and student models. \textbf{2) Mutual learning (Section \ref{Mutual learning})}.  The main idea of Probabilistic Teacher is to capture the uncertainty of unlabeled target data from a gradually evolving probabilistic teacher and guides the learning of a student in a mutually beneficial manner. To achieve this, based on Probabilistic Faster-RCNN, Probabilistic Teacher delivers the weakly-augmented images from target domain to the teacher model to obtain pseudo boxes, and notably, category and location of each pseudo box are in the form of general distribution over label space and four Gaussian distributions, respectively. These pseudo boxes are then used to train the student via Uncertainty-Guided Consistency Training for both classification and localization branches. The student transfers its learned knowledge to the teacher via exponential moving average (EMA). In this way, both models can evolve jointly and continuously to improve performance.

\subsection{Mutual Learning}
\label{Mutual learning}
\subsubsection{Uncertainty-Guided Consistency Training}
The student model is optimized on the labeled source data and the unlabeled target data with pseudo boxes generated from the teacher model. The training objective is written as:
\begin{equation}
\begin{small}
\begin{aligned}
\mathcal{L}_{total}=\mathcal{L}_{S}+\lambda_{T} \mathcal{L}_{T}
\end{aligned}
\end{small}
\end{equation}
where $\mathcal{L}_{S}$ is a supervised loss on labeled source data which is identical to Eqn.\ref{source-loss}. $\mathcal{L}_{T}$ is a self-supervised loss on unlabeled target data, which imposes the uncertainty-guided consistency between the teacher and student models. $\lambda_T$ is the loss weight for target domain, and is set to 1 by default.

To optimize the second term, target data with weak augmentation is fed into the teacher model to generate pseudo boxes, which contains classification probability distributions $p^{PL}$ and bbox coordinate probability distributions $t^{PL}$. Both distributions are sharpened to guide the student training. Specifically, $\mathcal{L}_{T}$ consists of four training losses, including two classification losses and two bbox regression losses in RPN and ROIhead:
\begin{equation}
\begin{small}
\begin{aligned}
\mathcal{L}_{T}=\mathcal{L}_{T-cls}^{RPN}+\mathcal{L}_{T-cls}^{ROI}+\mathcal{L}_{T-box}^{RPN}+\mathcal{L}_{T-box}^{ROI}
\end{aligned}
\end{small}
\end{equation}

The first two terms can be formulated as:
\begin{equation}
\begin{small}
\begin{aligned}
& \mathcal{L}_{T-cls}^{RPN}= \frac{1}{N_{cls}^{RPN}} \sum_{i} \mathcal{H}(\mathcal{M}(\mathcal{S}_{cls}(p_{i}^{PL}, \tau_{cls})), p_{i}^{RPN}) \\
&\mathcal{L}_{T-cls}^{ROI}=\frac{1}{N_{cls}^{ROI}} \sum_{i} \mathcal{H}( \mathcal{S}_{cls}(p_{i}^{PL}, \tau_{cls}), p_{i}^{ROI}) 
\end{aligned}
\end{small}
\end{equation}
where $p_{i}^{PL}$ is the $i_{th}$ classification probability distribution predicted by the teacher. $p_{i}^{RPN}$ and $p_{i}^{ROI}$ are the $i_{th}$ classification probability distribution in RPN and ROIhead predicted by the student. $\mathcal{S}_{cls}(\cdot, \tau_{cls})$ is a sharpening function where $\tau_{cls}$ is a temperature factor, which will be introduced later. $\mathcal{M}(\cdot)$ is a merging operation to sum up all foreground category probabilities to achieve the foreground/background probability distribution to guide RPN training. $N_{cls}^{RPN}$ and $N_{cls}^{ROI}$ are the batch size in RPN and ROIhead respectively. 

The last two terms can be unified into a general form:
\begin{equation}
\begin{small}
\begin{aligned}
\mathcal{L}_{T-box} =\frac{1}{N_{bbox}} \sum_{i} \mathbb{I}_{fg}(t_{i}) \mathcal{H}(\mathcal{S}_{bbox}(t_{i}^{PL}, \tau_{bbox} ), t_{i})
\label{equation_T_box}
\end{aligned}
\end{small}
\end{equation}
where $t_{i}^{PL}$ is the $i_{th}$ bbox coordinate probability distribution predicted by the teacher (Note that both terms are predicted in ROIhead). $t_{i}$ is the bbox coordinate probability distribution predicted by the student in RPN or ROIhead and associated with $t_{i}^{PL}$. $\mathcal{S}_{bbox}(\cdot ,\tau_{bbox})$ is a sharpening function for bbox regression and $\tau_{bbox}$ is a temperature factor.

\begin{figure*}[t]
        \centering
        \includegraphics[width=0.9\textwidth]{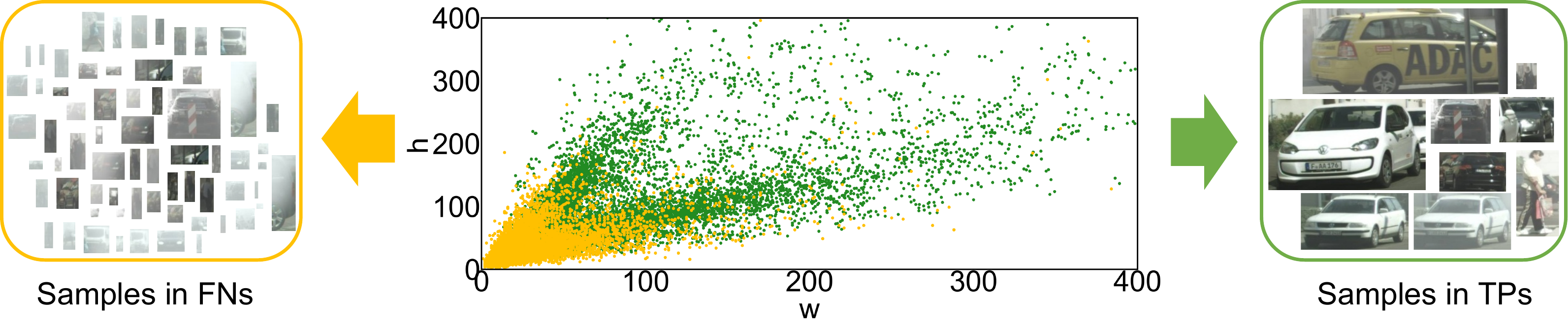}
        \vskip -0.15in
        \caption{Intra-domain gap in the ``Cityscapes to Foggy Cityscapes'' adaptation task. True positives (TPs, in \textbf{\textcolor{ForestGreen}{green}} color) and false negatives (FNs, in \textbf{\textcolor{Dandelion}{gold}} color) predicted by the ``source only'' model on the target domain. ``h'' and ``w'' represent the height and width of the predicted bounding boxes, respectively.}
        \label{fig_aug}
        \vskip -0.05in
\end{figure*}

\subsubsection{Sharpening Functions} 
Sharpening functions encourage the predictions to be sharp and low-entropy.
\begin{itemize}[leftmargin=12pt, topsep=2pt, itemsep=0pt]
\item For classification branch, $\mathcal{S}_{cls}(\cdot, \tau_{cls})$ is defined as a SoftMax function with temperature $\tau_{cls}$.
\begin{equation}
\begin{small}
\begin{aligned}
\mathcal{S}_{cls}(\cdot, \tau_{cls}) = SoftMax(\cdot, \tau_{cls})
\end{aligned}
\end{small}
\end{equation} 

\item For bbox regression branch, the entropy of Gaussian distribution is a function of its variance ${\sigma}^{2}$ (see the appendix \ref{appendix_entropy_Gaussian} for the proof). Hence, $\mathcal{S}_{bbox}(\cdot ,\tau_{bbox})$ is designed as:
\begin{equation}
\begin{small}
\begin{aligned}
\sigma^{2} \leftarrow \sigma^{2} * \tau_{bbox}
\end{aligned}
\end{small}
\end{equation}
\end{itemize}

When $\tau=1$ (including $\tau_{cls}$ and $\tau_{bbox}$), sharpening function is equivalent to the original SoftMax or Gaussian function. When $\tau\rightarrow 0$ or $\tau\rightarrow +\infty$, it tends to be a Dirac delta distribution or a uniform distribution, which corresponds the most low-entropy or high-entropy case. It is set as $\tau<1$ in this work. With this specialized sharpening function for bbox regression, $\mathcal{L}_{T-bbox}$ in Eqn.\ref{equation_T_box} is detailed as:
\begin{footnotesize} 
\begin{equation}
\begin{small}
\begin{aligned}
\mathcal{L}_{T-bbox} = & \frac{1}{N_{bbox}} \sum_{i} \mathbb{I}_{fg}(t_{i}) [ \log(\sigma_i) \\
& + \frac{ (\sigma_i^{PL})^2 *\tau_{bbox} + (\mu_i^{PL} - \mu_i)^2}{ \sigma_i^2} ] + C \\
\end{aligned}
\end{small}
\end{equation}
\end{footnotesize}where ($\mu_i^{PL}$, $\sigma_i^{P L}$) and ($\mu_i$, $\sigma_i$) are the $i_{th}$ mean and variance of bbox coordinates predicted by the teacher and student models, respectively. $C$ is a constant. See the appendix \ref{appendix_two_Gaussian} for the detailed proof of this step.

\subsubsection{Teacher Updating}
To obtain more accurate pseudo boxes, EMA is applied to gradually update the teacher model with positive feedbacks from the student model. Given the weight of the student $\theta^{S}$, the teacher $\theta^{T}$ is obtained by : $ \theta^{T}=\alpha \theta^{T}+(1-\alpha) \theta^{S}$, where $\alpha$ is the EMA rate. The slowly updated teacher model is regarded as an ensemble of student models in different training timestamps.

\renewcommand\arraystretch{1}
\begin{table*}[th]
	\centering
	\vskip -0.1in
	\caption{Results of adaptation from normal to foggy weather (C2F). ``\dag'' represents that the results are reproduced by us using the released codes, with the same strong augmentation and Probabilistic Faster-RCNN. ``Source only'' and  ``Oracle'' refer to the models trained by only using labeled source data and labeled target data, respectively. ``FR'' represents Faster-RCNN. ``0.01'', ``0.02'' and ``ALL'' in the column of ``split'' represent the foggy level 0.01, 0.02 and all three foggy levels, respectively. ``UN'' means that the foggy level used in the corresponding paper is unknown based on the paper and released code (if available).}
	\begin{threeparttable}
	\resizebox{.99\textwidth}{!}{
	\begin{tabular}{l | c | c | c | cccccccc | c}
		\toprule
		Methods & Split & Reference & Arch. & truck  & car & rider & person & train & motor & bicycle & bus  & mAP\\
		\midrule
		Source only & 0.02 & - & \multirow{8}{*}{FR+VGG16} & 9.0 & 28.5 & 26.6 & 22.4 & 	4.3 & 15.2 & 25.3 & 16.0  & 18.4 \\
		Source only (strong aug.) & 0.02  & - &  & 10.1 & 37.7 & 33.0 & 26.4 & 6.2 & 17.4 & 30.2 & 24.4 & 23.2\\
		Source only & ALL  & - &  & 12.1 & 40.4 & 33.4  & 27.9 & 10.1 & 20.7 & 30.9 & 23.2 & 24.8 \\
		Source only (strong aug.) & ALL  & - & & 18.6 & 48.1 & 41.3 & 33.9 & 7.3 & 26.6 & 37.8 & 34.5 & 31.0  \\
		\cline{1-3}
		\cline{5-13}
		MTOR \cite{EOR} & 0.02  & CVPR'19 & & 21.9 & 44.0 & 41.4 & 30.6 & 40.2 & 31.7 & 33.2 & 43.4 & 35.1 \\
		SW \cite{4} & 0.02  & CVPR'19  & & 24.5 & 43.5 & 42.3 & 29.9 & 32.6 & 30.0 & 35.3 & 32.6 & 34.3 \\
		DM \cite{9} & UN & CVPR'19 & & 27.2 & 40.5 & 40.5 & 30.8 & 34.5 & 28.4 & 32.3 & 38.4 & 34.6 \\
		PDA \cite{8} & ALL & WACV'20 & & 24.3 & 54.4 & 45.5 & 36.0 & 25.8 & 29.1 & 35.9 & 44.1  & 36.9\\
		\cline{4-4}
		GPA \cite{GPA} & 0.01 & CVPR'20 & FR+Resnet50 & 24.7 & 54.1 & 46.7 & 32.9 & 41.1 & 32.4 & 38.7 & 45.7  & 39.5 \\
		\cline{4-4}
		ATF \cite{ATF} & UN & ECCV'20 & \multirow{16}{*}{FR+VGG16}  & 23.7 & 50.0 & 47.0 & 34.6 & 38.7 & 33.4 & 38.8 & 43.3 & 38.7  \\
		HTCN \cite{HTCN} & 0.02  & CVPR'20  & & 31.6 & 47.9 & 47.5 & 33.2 & 40.9 & 32.3 & 37.1 & 47.4 & 39.8 \\
		ICR-CCR \cite{ICR-CCR} & ALL  & CVPR'20 & & 27.2 & 49.2 & 43.8 & 32.9 & 36.4 & 30.3 & 34.6 & 36.4 & 37.4 \\
		CF \cite{coarse} & UN & CVPR'20 & & 30.8 & 52.1 & 46.9 & 34.0 & 29.9 & 34.7 & 37.4 & 43.2  & 38.6 \\
		iFAN \cite{ifan} & UN & AAAI'20 &  & 27.9 & 48.5 & 40.0 & 32.6 & 31.7 & 22.8 & 33.0 & 45.5  & 35.3 \\
		SFOD \cite{SFOD} & ALL & AAAI'21 & & 25.5 & 44.5 & 40.7 & 33.2 & 22.2 & 28.4 & 34.1 & 39.0  & 33.5 \\
		MeGA \cite{mega} & UN  & CVPR'21 & & 25.4 & 52.4 & 49.0 & 37.7 & \textbf{46.9} & 34.5 & 39.0 & 49.2  & 41.8\\
		UMT  \cite{10} & 0.02 & CVPR'21 & & \textbf{34.1} & 48.6 & 46.7 & 33.0 & 46.8 & 30.4 & 37.3 & 56.5 & 41.7  \\
		\cline{1-3}
		\cline{5-13}
		SW \dag \cite{4} & ALL & CVPR'19 & & 28.7 & 51.0 & 46.3 & 34.2  & 24.0 & 33.8 & 37.1 & 44.9  & 37.5 \\
		ICR-CCR \dag \cite{ICR-CCR} & ALL & CVPR'20 & & 29.7 & 50.4  & 47.2 & 33.6 & 35.1 & 34.6 & 37.9 & 50.0 & 39.8 \\ 
		\cline{1-3}
		\cline{5-13}
		PT & 0.02 & Ours & & 30.7 & 59.7 & 48.8 & 40.2 & 30.6 & 35.4 & 44.5 & 51.8  & 42.7 \\
		PT & ALL & Ours & & 33.4 & \textbf{63.4} & \textbf{52.4} & \textbf{43.2} & 37.8 & \textbf{41.3} & \textbf{48.7} & \textbf{56.6}  & \textbf{47.1}  \\
		\cline{1-3}
		\cline{5-13}
		Oracle & 0.02 & - & & 33.1 & 59.1 & 47.3 & 39.5 & 42.9 & 38.1 & 40.8 & 47.3  & 43.5\\
		Oracle & ALL & - & & 32.6 & 61.6 & 49.1 & 41.2 & 49.0 & 37.9 & 42.4  &  56.6  & 46.3  \\
	 \bottomrule
	\end{tabular}
	}
	\end{threeparttable}
	\label{tab:c2f}
	\vspace{-1em}
\end{table*}

\subsubsection{Anchor Adaptation}
In the existing works, source and target domains usually share the same anchors. However, source and target domains generally have different distributions of bbox sizes due to domain shifts. It is intuitive to adapt the anchors automatically during self-training. 
Specifically, anchors have been proven to be learnable parameters \cite{ABO}. We propose to adapt the anchor shapes slowly during teacher-student mutual learning in an EMA mechanism to match the distribution of bboxes in target domain. The overall optimization objective is:

\begin{equation}
\begin{small}
\begin{aligned}
\min _{\theta^{S}} \{ {\mathcal{L}_{S} + \min _{ \mathrm{\{(w_k, h_k)\}_{k=1}^{A}}} {\lambda_{T} \mathcal{L}_{T}}} \}
\end{aligned}
\end{small}
\end{equation}

where $\{(w_k, h_k)\}_{k=1}^{A}$ is the anchor shapes, and \textit{A} is the number of anchors.

\subsection{Entropy Focal Loss}
Although PT has exploited the noisy pseudo boxes in a step-by-step manner that the predictions are encouraged to be low-entropy gradually, the existing noisy pseudo boxes may inevitably harm the performance. Since the proposed framework can obtain the uncertainty of each bbox (category plus four coordinates), it is reasonable to apply these uncertainty information to improve performance. Based on this intuitive thought, we use the entropy of category and location to describe the uncertainty of each bbox, and introduce an entropy focal loss to further facilitate uncertainty-guided consistency training for both classification and localization branches. Entropy focal loss for classification and regression branches can be unified into a general form:
\begin{equation}
\begin{small}
\begin{aligned}
\mathcal{H}_{EFL}(\cdot, \cdot)=\left(1-\mathcal{E} / \mathcal{E}_{norm}\right)^{\lambda} \mathcal{H}(\cdot, \cdot)
\end{aligned}
\end{small}
\end{equation}
where $\lambda$ is a hyperparameter, $\mathcal{E}$ is the prediction entropy from the teacher, and $\mathcal{E}_{norm}$ is the norm term. $\mathcal{E}_{norm}$ is set as the maximal value of the entropy. Theoretically, $\mathcal{E}_{norm}$ equals $\log(n+1)$ for classification and equals $\frac{1}{2}\log(2 \pi)+\frac{1}{2}$ for localization (see the \textbf{Appendix} \ref{appendix_entropy_Gaussian} and \ref{appendix_Maximal_entropy} for the detailed proof), where $n$ is the number of foreground classes.

With the obtained uncertainty for the category and each coordinate of pseudo boxes, Entropy Focal Loss encourages the model to pay more attention to less noisy category predictions in classification branch and more accurate coordinate individuals in regression branch.

\section{Strong Augmentation for UDA-OD}
\label{Intra_Domain_Gap}

\subsection{Intra-Domain Gap}
\label{Intra-Domain}
We visualize the true positives (TPs) and false negatives (FNs) predicted by the ``source only'' model on the target domain in Fig.\ref{fig_aug}. 
We obeserve that the smaller, severelier blurred and occluded objects tend to have a poorer adaptation performance and vice verse. We dub this phenomenon as \emph{intra-domain gap}. 
As observed in qualitative results in Fig.\ref{fig_visualization}, intra-domain gap are commonly-exist in UDA-OD community. 
However, unlike inter-domain gap has been addressed in many previous works, intra-domain gap, one of the bottlenecks restricting the performance of UDA-OD, in contrast, has been neglected.

\subsection{Intra-Domain Alignment via Strong Augmentation}
Our findings about the issue of intra-domain gap in UDA-OD leads us to introduce strong data augmentation into our framework.  
Concretely, since large-scale, distinct objects usually attain high confidence score during pseudo labeling, 
we transform these to mimic those small scale, blurred and occluded ones via strong data augmentation (random resizing, guassian blur, color jitter and etc.). 
In this way, these transformed objects with low-entropy pseudo labels will guide the model to pay more attention to small scale, blurred and occluded ones. 
From this perspective, strong data augmentation is actually an implicit intra-domain alignment method to bridge the intra-domain gap.

\section{Experiments}
\subsection{Experimental Settings}
\textbf{Datasets}.
To validate our approach, we conduct extensive experiments on multiple benchmarks with four different types of domain shifts, including 1) C2F: adaptation from normal to foggy weather, 2) C2B: adaptation from small to large-scale dataset, 3) K2C: adaptation across cameras, 4) S2C: adaptation from synthetic to real images. Five public datasets are used in our experiments.

\begin{itemize}[leftmargin=12pt, topsep=2pt, itemsep=0pt]
\item \textbf{Cityscapes (C)} \cite{18} contains 2,975 training images and 500 validation images with pixel-level annotations. The annotations are transformed into bounding boxes for the following experiments. 
\item \textbf{Foggy Cityscapes (F)} \cite{19} is a synthetic dataset rendered from Cityscapes with three levels of foggy weather (0.005, 0.01, 0.02), which correspond to the visibility ranges of 600, 300 and 150 meters. 
\item \textbf{BDD100k (B)} \cite{20} is a large-scale dataset consisting of 100k images. The subset of images labeled as daytime, including 36,728 training and 5,258 validation images, are used for the following experiments. 
\item \textbf{Sim10k (S)} \cite{21} consists of 10k images rendered by a gaming engine. In Sim10k, bounding boxes of 58,701 cars are provided in the 10,000 training images. All images are used in the experiments.
\item \textbf{KITTI (K)} \cite{22} is collected by an autonomous driving platform, including 14999 images and 80256 bounding boxes. Only the train set is used here.
\end{itemize}

\textbf{Network Architecture}. 
We take Faster-RCNN as the base detector following \cite{3}, where VGG16 \cite{24} pre-trained on ImageNet \cite{25} is used as its backbone. We rescale all images by setting the shorter side of each image to 600 while keeping the aspect ratios unchanged. Our implementation is built upon Detectron2 \cite{27}.

\renewcommand\arraystretch{1}
\begin{table}[t]
	\centering
	\vskip -0.1in
	\caption{Results of adaptation from small to large-scale dataset (C2B). Please refer to Table \ref{tab:c2f} for notations illustration.}
	\begin{threeparttable}
	\resizebox{.48\textwidth}{!}{
	\begin{tabular}{l | c | c | c}
		\toprule
		Methods & Reference & Arch.& mAP \\
		\midrule
		Source only & -  & \multirow{10}{*}{FR+VGG16} & 20.6 \\
		Source only (strong aug.) & -  &  & 26.9  \\	
		\cline{1-2} \cline{4-4}
		ICR-CCR \cite{ICR-CCR} & CVPR'20 &  & 26.9  \\
		SFOD \cite{SFOD} & AAAI'21  &  & 29.0  \\
		\cline{1-2} \cline{4-4}
		SW \dag  \cite{4} & CVPR'19  &  & 27.6 \\ 
		ICR-CCR  \dag \cite{ICR-CCR} & CVPR'20  &  & 29.5  \\ 
		\cline{1-2} \cline{4-4}
		PT  & Ours  &  &  \textbf{34.9}\\			
		Oracle & -  &  & 51.7 \\
	 \bottomrule
	\end{tabular}
	}
	\end{threeparttable}
	\label{tab:c2b}
	\vspace{-1em}
\end{table}

\renewcommand\arraystretch{1}
\begin{table}[t]
	\centering
	\vskip -0.1in
	\caption{Results of adaptation across cameras (K2C). Please refer to Table \ref{tab:c2f} for notations illustration.}
	\begin{threeparttable}
	\resizebox{.48\textwidth}{!}{
	\begin{tabular}{l | c | c | c }
		\toprule
		Methods &  Reference & Arch. & AP of car\\
		\midrule
		Source only & - & \multirow{3}{*}{FR+VGG16} & 40.3  \\
		Source only (strong aug.) & - & & 46.4  \\
		\cline{1-2}
		\cline{4-4}
		ATF \cite{ATF} & ECCV'20 & & 42.1  \\
		\cline{3-3}
		GPA \cite{GPA} & CVPR'20 & FR+Resnet50 & 47.9 \\
		\cline{3-3}
		SFOD \cite{SFOD} & AAAI'21 & \multirow{2}{*}{FR+VGG16} & 44.6  \\
		MeGA \cite{mega} & CVPR'21 & & 43.0  \\
		\cline{3-3}
		SimROD  \cite{11} & ICCV'21 & YOLOv5 & 47.5  \\
		\midrule
		SW \dag \cite{4} & CVPR'19 &  \multirow{6}{*}{FR+VGG16}  & 47.1  \\
		ICR-CCR \dag \cite{ICR-CCR} & CVPR'20 & & 47.6  \\
		\cline{1-2}
		\cline{4-4}
		PT & Ours & & \textbf{60.2}  \\
		Oracle & - & & 66.4  \\
	 \bottomrule
	\end{tabular}
	}
	\end{threeparttable}
	\label{tab:k2c}
	\vspace{-1em}
\end{table}

\textbf{Strong Augmentation}. 
We use the same data augmentation strategy in \cite{28} except RandomResizedCrop. Instead, random resizing is applied to the images. Weak augmentation refers to random horizontal-flipping. See Appendix \ref{more_implementation_details} for more implementation details.

\textbf{Optimization}. 
We use a batch size of 16 for both source and target data on a single GPU and train for 30k iterations with a fixed learning rate of 0.016, including 4k iterations for Pretraining and 26k iterations for Multual Learning. The detector is trained by an SGD optimizer with the momentum of 0.9 and the weight decay of $10^{-4}$. The EMA rate $\alpha$ is set to 0.9996. Without careful tuning, the loss weights in this paper are all set to 1. Moreover, $\lambda$ in EFL, together with temperature $\tau_{cls}$ and $\tau_{bbox}$, are all set to 0.5 simply.

\textbf{Evaluation Protocol and Comparison Baselines}. 
Following the existing works, we evaluate our method with the standard mean average precision (mAP) at the IOU threshold of 0.5. In our experiments, we observe that both strong augmentation and Probabilistic Faster-RCNN contribute to the baseline performance. For a fair comparison, first, we reproduce two existing works SW~\cite{4} and ICR-CCR~\cite{ICR-CCR}, using released codes with strong augmentation and Probabilistic Faster-RCNN. Second, we conduct extensive ablation studies in Section \ref{ablation study}.

\renewcommand\arraystretch{1} 
\begin{table}[t]
	\centering
	\vskip -0.1in
	\caption{Results of adaptation from synthetic to real images (S2C). Please refer to Table \ref{tab:c2f} for notations illustration.}
	\begin{threeparttable}
	\resizebox{.48\textwidth}{!}{
	\begin{tabular}{l | c | c | c }
		\toprule
		Methods & Reference & Arch. & AP of car \\
		\midrule
		Source only  & - & \multirow{7}{*}{FR+VGG16} & 35.5 \\
		Source only (strong aug.)  & - & & 44.5 \\
		\cline{1-2}
		\cline{4-4}
		SW \cite{4}  & CVPR'19 & & 40.7 \\
		ATF \cite{ATF} & ECCV'20 & & 42.8 \\
		HTCN  \cite{HTCN}  & CVPR'20 & & 42.5 \\
		iFAN  \cite{ifan}  & AAAI'20 &  & 46.9 \\
		CF \cite{coarse} & CVPR'20 & & 43.8 \\
		\cline{3-3}
		GPA \cite{GPA}  & CVPR'20 & FR+Resnet50  & 47.6 \\
		\cline{3-3}
		SFOD \cite{SFOD} & AAAI'21 & \multirow{3}{*}{FR+VGG16} & 42.9 \\
		MeGA \cite{mega} & CVPR'21 & & 44.8 \\
		UMT \cite{10} & CVPR'21 & & 43.1 \\
		\cline{3-3}
		SimROD \cite{11} & ICCV'21 & YOLOv5 & 52.1 \\
		\midrule
		SW \dag \cite{4} & CVPR'19 & \multirow{6}{*}{FR+VGG16}& 45.4  \\
		ICR-CCR \dag \cite{ICR-CCR} & CVPR'20 & & 46.1  \\
		\cline{1-2}
		\cline{4-4}
		PT & Ours & & \textbf{55.1}  \\
		Oracle & - &  & 66.4  \\
	 \bottomrule
	\end{tabular}
	}
	\end{threeparttable}
	\label{tab:s2c}
	\vspace{-1em}
\end{table}

\subsection{C2F: Adaptation from Normal to Foggy Weather}
In real-world scenarios, such as automatic driving, object detectors may be applied under different weather conditions. To study adaptation from normal to foggy weather, we use labeled Cityscapes and unlabeled Foggy Cityscapes (train set) for cross-domain self-training, and then report the evaluation results on the validation set of Foggy Cityscapes.

Notably, after investigating the existing works and the released code (if available), we find that the dataset spilts of Foggy Cityscapes they used differ from each other. We validate our approach on two most commonly used splits, level 0.02 and all three levels for a fair comparison. We use all three levels in the following ablation studies. Table \ref{tab:c2f} shows that our method achieves a new SOTA result in both splits. Notably, strong augmentation improves ``source only'' of level 0.02 and all three levels by +4.8 and +6.2 mAP, respectively. Compared with other approaches, our method achieves 42.7 mAP for level 0.02 and 47.1 mAP for all three levels, outperforming the best baseline MeGA \cite{mega} by a large margin (+0.9/+5.3).

\renewcommand\arraystretch{1}
\begin{table}[t]
	\centering
	\vskip -0.1in
	\caption{The effectiveness of anchor adaptation (AA) and EFL.}
	\begin{threeparttable}
	\resizebox{0.48\textwidth}{!}{
	\begin{tabular}{l | c | c | c | c }
		\toprule 
		Methods & C2F & C2B & K2C & S2C \\
		\midrule
		PT & 47.1 & 34.9 & 60.2 & 55.1 \\
		PT w/o AA & 46.5 \textcolor{blue}{(-0.6)}  & 34.4 \textcolor{blue}{(-0.5)} & 59.0 \textcolor{blue}{(-1.2)} & 54.3 \textcolor{blue}{(-0.8)} \\
		PT w/o AA \& EFL & 45.4 \textcolor{blue}{(-1.7)} & 32.9 \textcolor{blue}{(-2.0)} & 58.6 \textcolor{blue}{(-1.6)} & 54.1 \textcolor{blue}{(-1.0)} \\						
	 \bottomrule
	\end{tabular}
	}
	\end{threeparttable}
	\label{tab:aa-efl}
	\vspace{-1.5em}
\end{table}

\renewcommand\arraystretch{1}
\begin{table}[t]
	\centering
	\vskip -0.1in
	\caption{Results of extension to source-free setting.}
	\begin{threeparttable}
	\resizebox{0.48\textwidth}{!}{
	\begin{tabular}{l | c | c | c | c | c | c}
		\toprule 
		\multirow{2}{*}{Methods} & \multirow{2}{*}{Source data} & \multirow{2}{*}{Reference} & \multicolumn{4}{c}{mAP} \\
		\cline{4-7} \\[-1.5ex]
		&  &  & C2F & C2B & K2C & S2C \\
		\midrule
		SFOD & \XSolidBrush & AAAI'21 & 33.5 & 29.0 &  44.6 & 42.9 \\
		\midrule
		PT & \XSolidBrush  & Ours & 38.7 &  28.1 & 59.6 & 54.1 \\
		PT & \Checkmark & Ours & \textbf{47.1} & \textbf{34.9} &  \textbf{60.2} & \textbf{55.1} \\						
		Oracle & - & - & 46.3 &  51.7 &  66.4 &  66.4  \\
	 \bottomrule
	\end{tabular}
	}
	\end{threeparttable}
	\label{tab:c2f-source-free}
	\vspace{-2em}
\end{table}

\subsection{C2B: Adaptation from Small to Large-Scale Dataset}
Currently, collecting and labeling large amounts of image data with different scene layout can be extremely costly, e.g., automatic driving from one city to another. To study the effectiveness of our method for adaptation to a large-scale dataset with different scene layout, we use Cityscapes as a smaller source domain dataset, BDD100k containing distinct attributes as a large unlabeled target domain dataset. Following ICR-CCR \cite{ICR-CCR} and SFOD \cite{SFOD}, we report the results on seven common categories on both datasets. Table \ref{tab:c2b} shows the results of this experiment, where our method outperforms all the baselines. Our method achieves 34.9 mAP, a large margin (+5.4) compared to ICR-CCR \cite{ICR-CCR}.

\subsection{K2C: Adaptation across Cameras}
Different camera setups (e.g., angle, resolution, quality, and type) widely exist in the real world, which causes domain shift. In this experiment, we study the adaptation between two real datasets. The KITTI and Cityscapes datasets are used as source and target domains, respectively. The results are provided in Table \ref{tab:k2c}. Our proposed method improves upon the best method GPA \cite{GPA} by +12.7 mAP. Notably, our method also outperforms the very recent work SimROD, which takes YOLOv5 \cite{YOLOv5} as the base detector and relies on a large-scale teacher model.

\subsection{S2C: Adaptation from Synthetic to Real Images}
Synthetic images offer an alternative to alleviate the data collection and annotation problems. However, there is a distribution gap between synthetic data and real data. To adapt the synthetic scenes to the real one, we utilize the entire Sim10k dataset as source data and the training set of Cityscapes as target data. Since only the \textit{car} category is annotated in both domains, we report the AP of car in the test set of Cityscapes. As provided in Table \ref{tab:s2c}, our method outperforms existing approaches by a large margin, which improves over the current best method by +3.0 mAP.

\begin{figure*}[htbp]
\centering
\vspace{-0.5em}
\subfigure[Source only]
{
    \begin{minipage}[b]{.3\linewidth}
        \centering
        \includegraphics[scale=0.09]{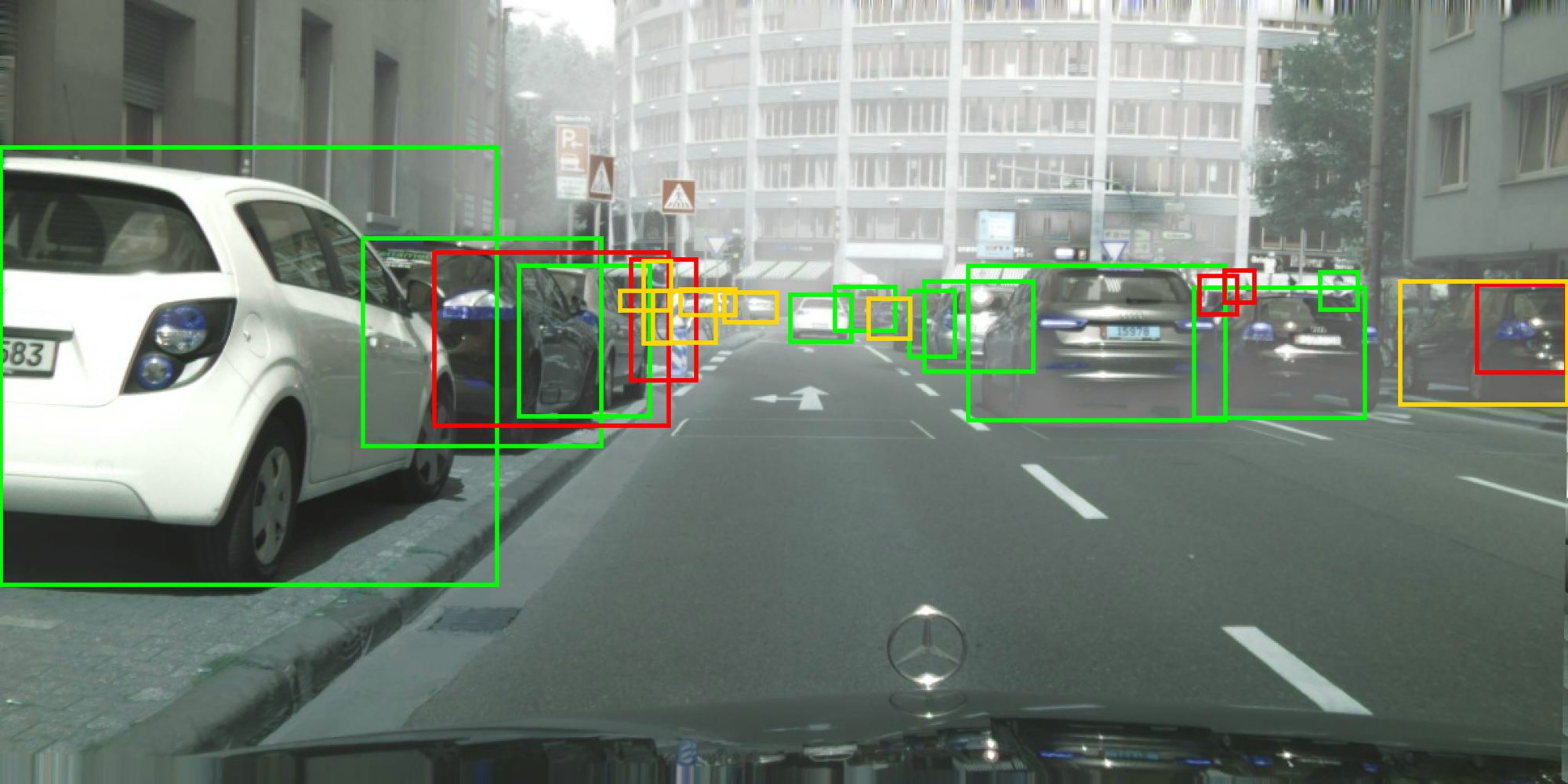} \\
        \includegraphics[scale=0.09]{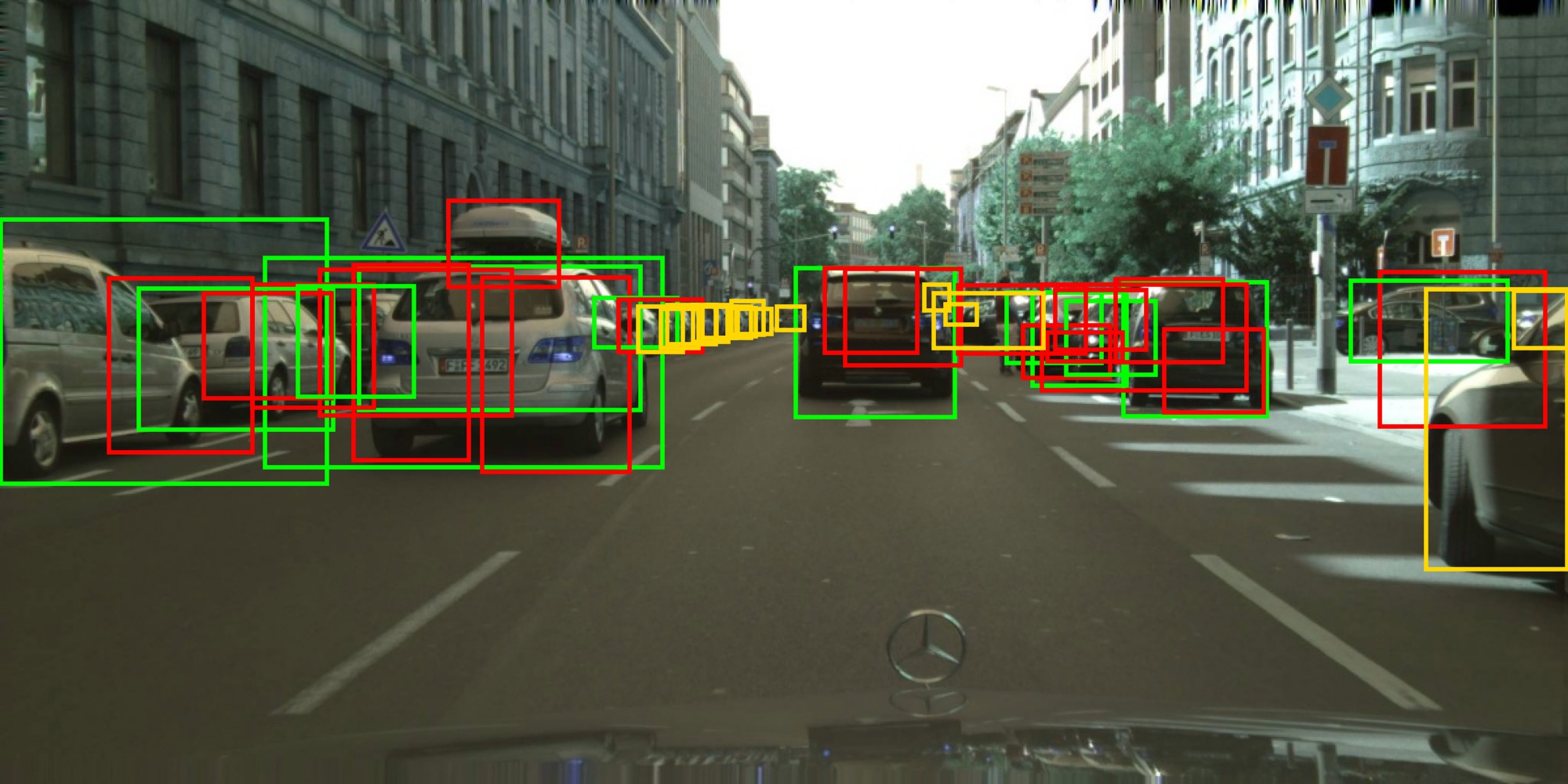}
    \end{minipage}
}
\centering
\subfigure[SW]
{
    \begin{minipage}[b]{.3\linewidth}
        \centering
        \includegraphics[scale=0.09]{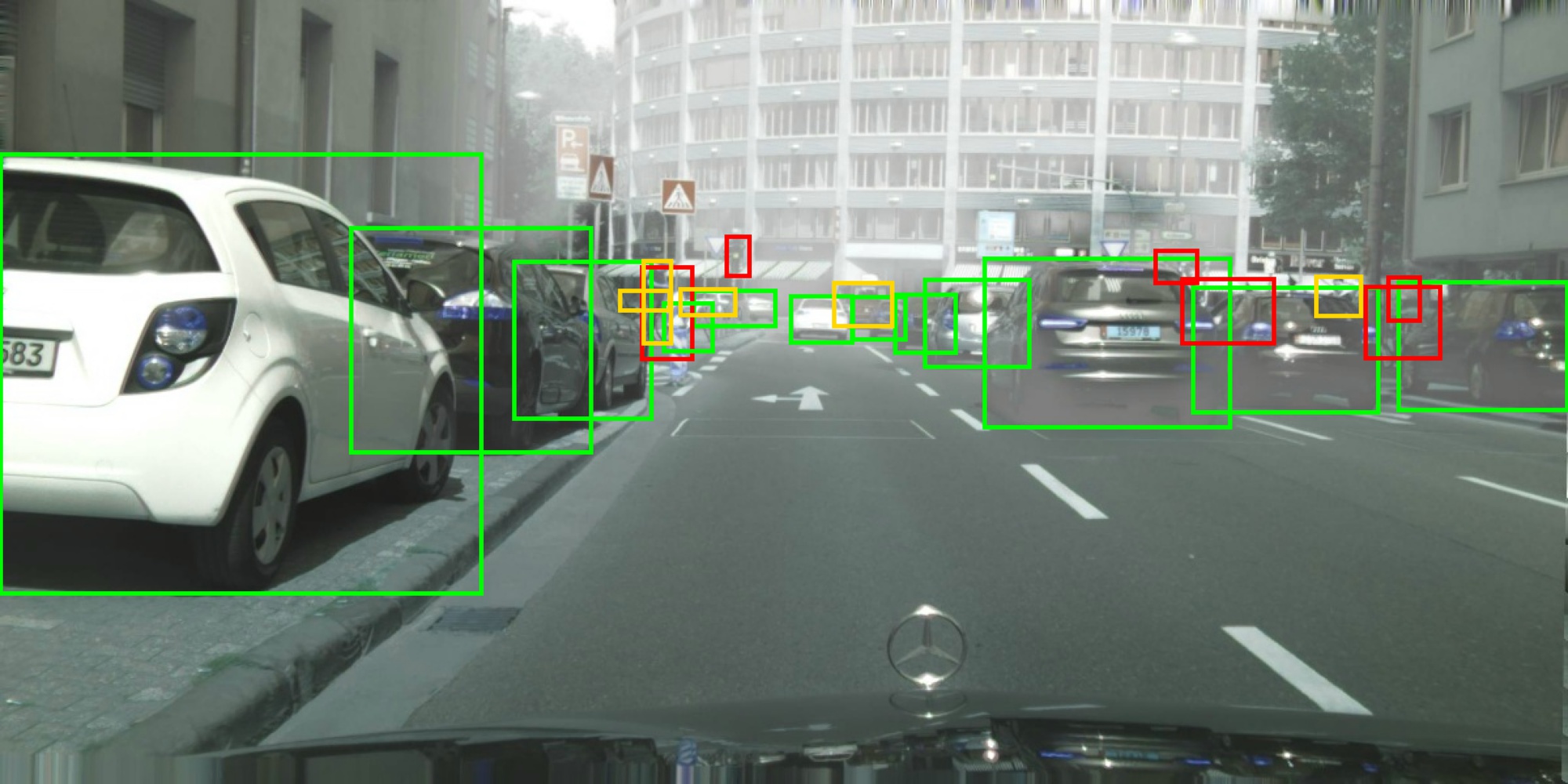} \\
        \includegraphics[scale=0.09]{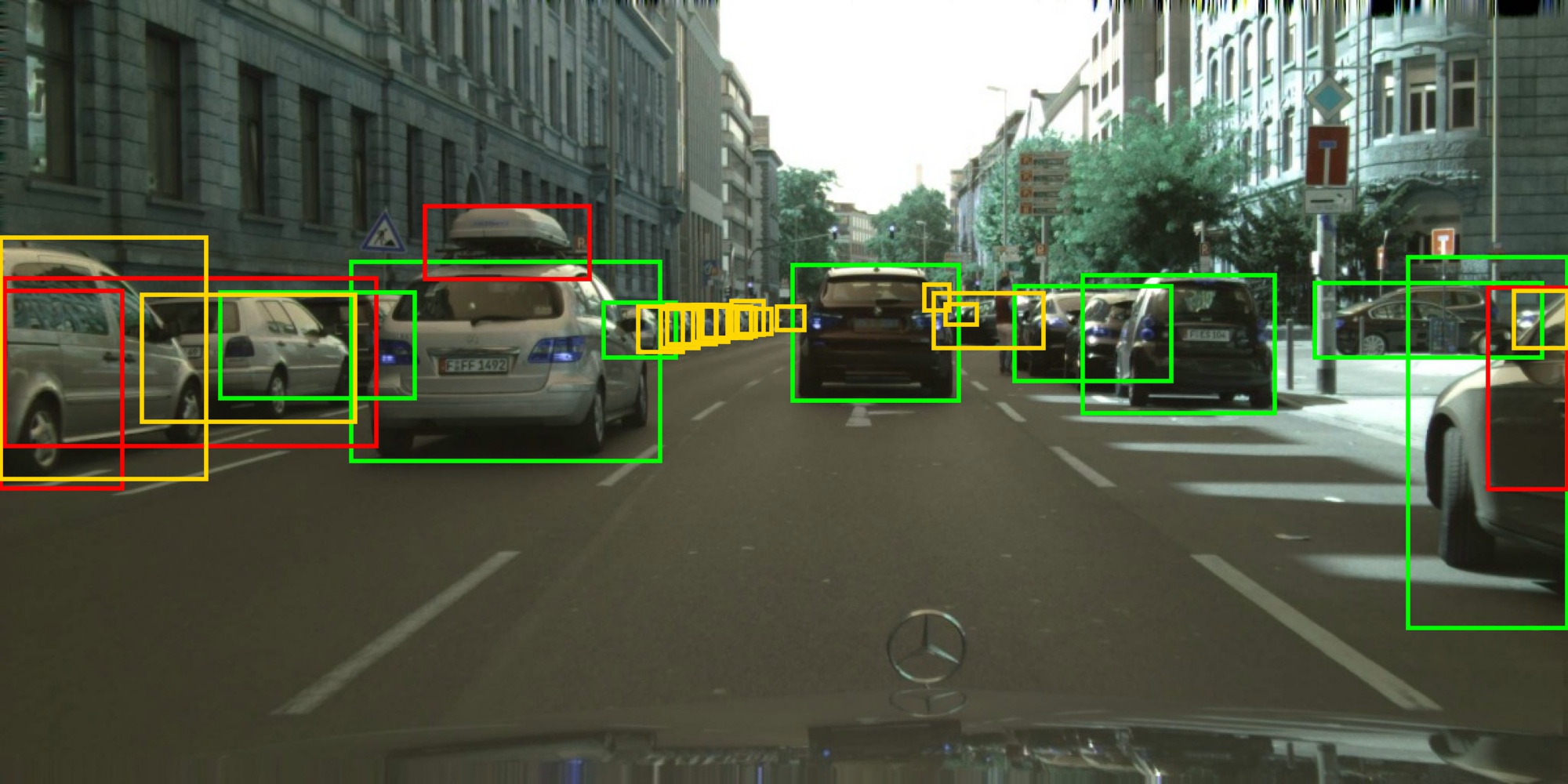}
    \end{minipage}
}
\centering
\subfigure[PT]
{
    \begin{minipage}[b]{.25\linewidth}
        \centering
        \includegraphics[scale=0.09]{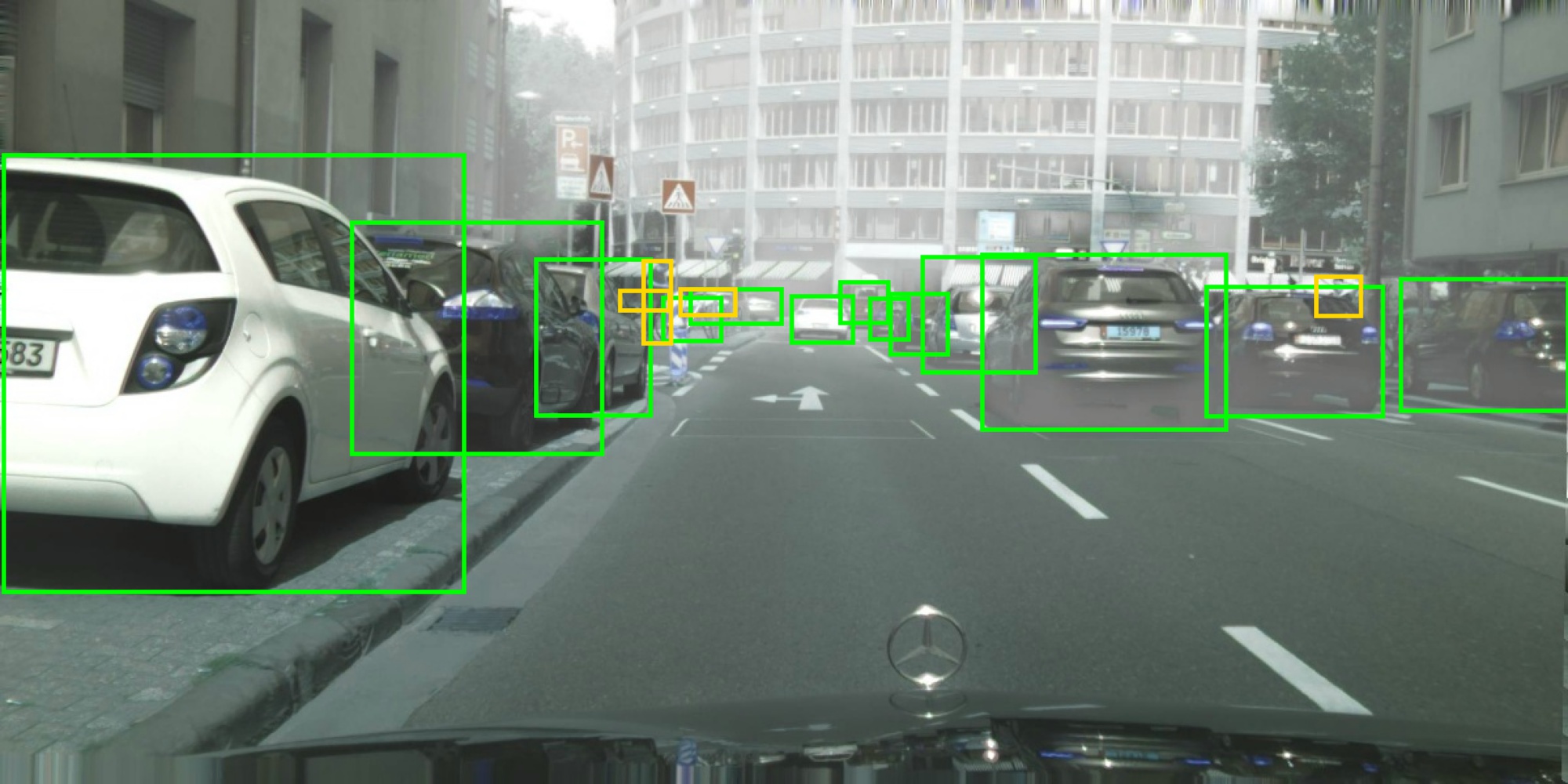} \\
        \includegraphics[scale=0.09]{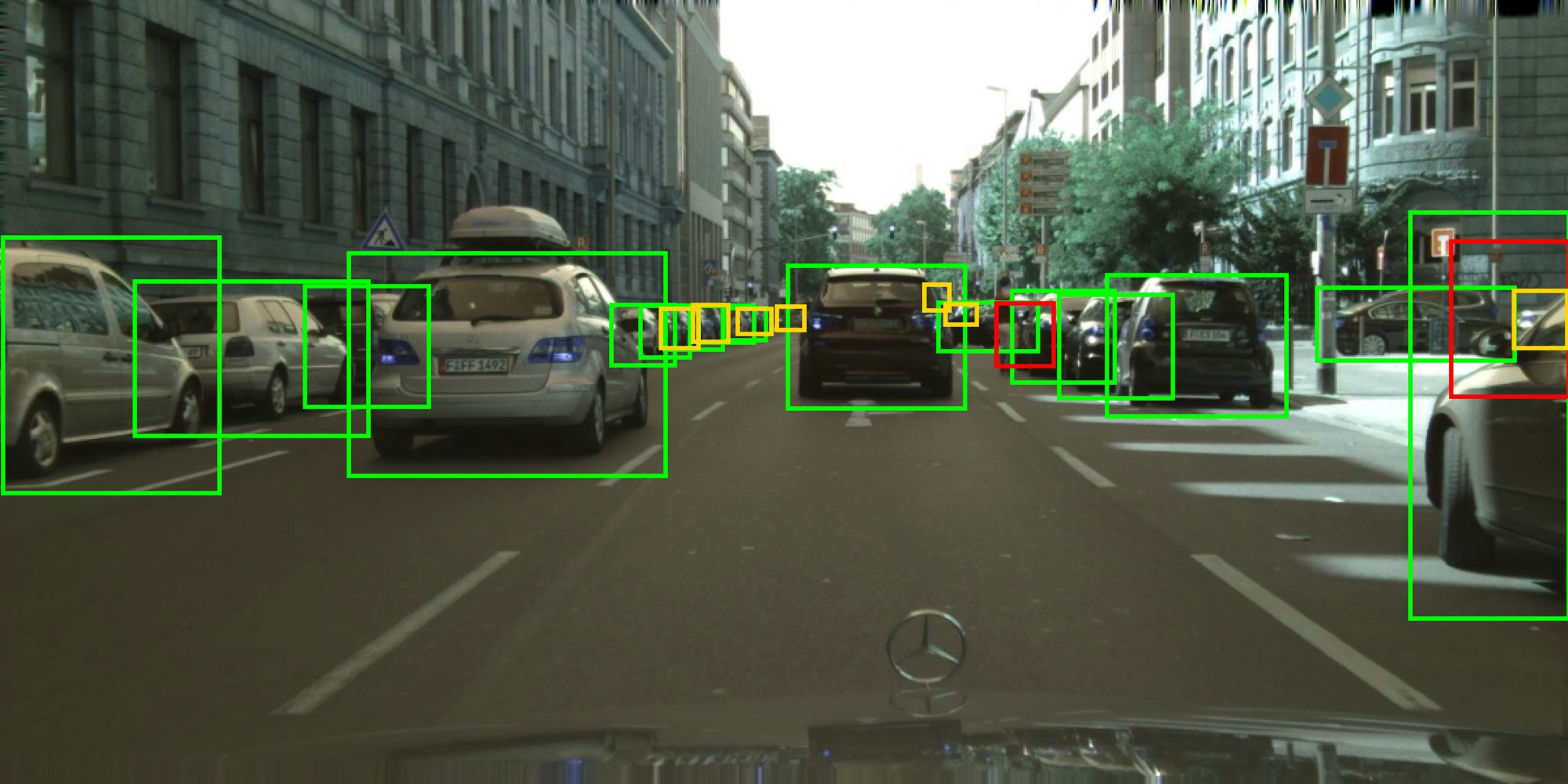}
    \end{minipage}
}
\vspace{-1em}
\caption{Qualitative results on the target domain of C2F (top row) and K2C (bottom row). \textbf{\textcolor{ForestGreen}{Green}}, \textbf{\textcolor{red}{red}} and \textbf{\textcolor{Dandelion}{gold}} boxes denote true positives, false positives and false negatives, respectively.}
\vspace{-1em}
\label{fig_visualization}
\end{figure*}

\subsection{Ablation Studies}
\label{ablation study}

\begin{wraptable}{r}{0.25\textwidth}
	\centering
	\vskip -0.7in
	\caption{Performance under different data augmentations. "F", "C", "G", "B", "S" and "R" denote Random Horizontal Flipping, Colorjitter, Grayscale, Gaussian Blur, Solarization and Resize, respectively.}
	\begin{threeparttable}
	\resizebox{0.25\textwidth}{!}{
	\begin{tabular}{c c c c c c | c}
		\toprule 
		F & C & G & B & S & R & mAP \\
		\midrule
		\Checkmark &  &  &  &  &  & 34.4  \\
		\Checkmark & \Checkmark &  &  &  &  & 37.6  \\
		\Checkmark & \Checkmark  & \Checkmark  &  &  &  & 38.5  \\
		\Checkmark & \Checkmark  & \Checkmark  & \Checkmark &  &  & 40.1  \\
		\Checkmark & \Checkmark  & \Checkmark  & \Checkmark & \Checkmark &  & 42.5  \\
		\Checkmark & \Checkmark  & \Checkmark  & \Checkmark & \Checkmark & \Checkmark & \textbf{47.1}  \\
	 \bottomrule
	\end{tabular}
	}
	\end{threeparttable}
	\label{tab:aug}
	\vskip -0.2in
\end{wraptable}

\textbf{Augmentation.}
Table \ref{tab:aug} shows that augmentation is crucial for learning domain adaptive object detection. Specifically, the stronger the augmentation is, the better the performance is. 
As claimed in Section \ref{Intra_Domain_Gap}, we argue that one serious issue neglected by previous works in UDA-OD is the intra-domain gap. Strong augmentation can be viewed as an implicit intra-domain alignment method to bridge the intra-domain gap between true labels and false labels in target domain, and that is why we introduce strong augmentation into our method. Considering this, as shown in Table 1-4, we have re-implemented two state-of-the-art domain alignment approaches with strong augmentation for fair comparisons, from which we find that strong data augmentation obtains much more performance enhancement in self-training paradigm than domain alignment counterparts.

\textbf{Anchor adaptation and EFL.}  
Table \ref{tab:aa-efl} shows the effectiveness of anchor adaptation (AA) and EFL, and it can be observed that both contribute to the improvement on all four different types of domain shifts. The effectiveness of AA indicates the necessity of addressing the objects size (scale) shifts in UDA-OD task, while the effectiveness of EFL supports the claim that the obtained entropy uncertainty of pseudo boxes can be used to further boost the performance. 

\textbf{Extension to source-free setting.}
Only with the self-supervised loss on target domain, PT can be seamlessly and effortlessly extended to source-free UDA-OD (privacy-critical scenario) \cite{SFOD,2021Box,SSNLL,TransductiveClip2022}, where only unlabeled target data is involved for the purpose of privacy protection. Table \ref{tab:c2f-source-free} shows that PT achieves substantial improvements, showing its robustness and scalability. Of particular interest, the performances of PT w/o and w/ source data in both K2C and S2C are almost comparable, while in C2F and C2B, there are still a large gap, which we remain for a future research.

\renewcommand\arraystretch{1}
\begin{table}[t]
	\centering
	\vskip -0.1in
	\caption{The effect of Probabilistic Faster-RCNN. ``Source only'' results are presented.}
	\begin{threeparttable}
	\resizebox{0.45\textwidth}{!}{
	\begin{tabular}{l | c | c | c | c }
		\toprule 
		Models & $\,\,$C2F$\,\,$ & $\,\,$C2B$\,\,$ & $\,\,$K2C$\,\,$ & $\,\,$S2C$\,\,$ \\
		\midrule
		Vanilla Faster-RCNN& 30.2  & 26.3 & 45.9 & 44.1 \\
		Probabilistic Faster-RCNN& 31.0 & 26.9 & 46.4 & 44.5 \\
	 \bottomrule
	\end{tabular}
	}
	\end{threeparttable}
	\label{Probabilistic}
	\vspace{-1em}
\end{table}

\renewcommand\arraystretch{1}
\begin{table}[t]
	\centering
    \caption{``Thres.'' and ``$\tau$'' are short for threshold and ($\tau_{cls}$, $\tau_{bbox}$).}
	\resizebox{0.48\textwidth}{!}{
	\begin{tabular}{c | c | c | c | c | c | c | c}
		\toprule 
		Thres. &  0.5 & 0.6 & 0.7 & 0.8 & 0.9 & \textbf{mean$\uparrow$} & \textbf{std$\downarrow$} \\
		\midrule
		mAP & 35.9 & 33.9 & 49.0 & 56.1 & 56.5 & \textbf{46.2} & \textbf{9.6} \\
		\midrule 
		\midrule
		$\tau$ & (0.25, 0.5) & (0.75, 0.5) & (0.5, 0.5) & (0.5, 0.25) & (0.5, 0.75) & \textbf{mean$\uparrow$} & \textbf{std$\downarrow$} \\
		\midrule
		mAP & 59.3 & 58.9 & 60.2 & 57.6 & 59.9 & \textbf{59.2} & \textbf{0.9} \\
	 \bottomrule
	\end{tabular}
	}
	\vspace{-1em}
	\label{comp_thre}
\end{table}

\begin{figure}[t]
    \centering
    \includegraphics[width=0.48\textwidth]{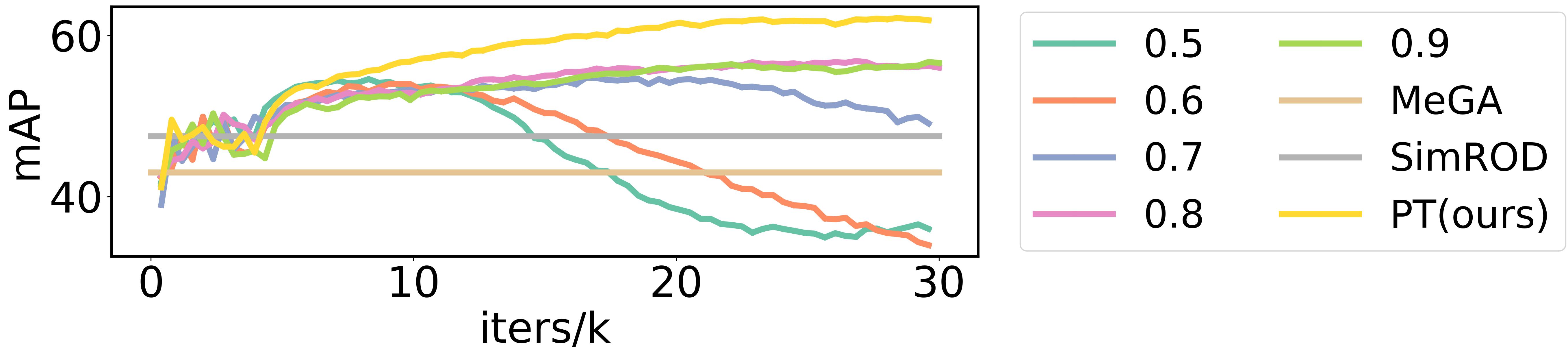}
    \vskip -0.1in
    \caption{Performance curve of K2C, using PT and vanilla Faster-RCNN with different confidence thresholds.}
    \label{reb}
    \vspace{-1.5em}
\end{figure}

\textbf{The effect of uncertainty.}
We perform the self-training of Faster-RCNN with different confidence thresholds under the same mean-teacher framework. As shown in Fig.\ref{reb}, we observe the same phenomenon as \cite{SFOD} that the performance varies widely across different thresholds. The proposed PT, in contrast, a threshold-free approach, achieves remarkable SOTA results compared to Faster-RCNN with different confidence thresholds.

\textbf{The effect of probabilistic modeling.}
PT is derived from Probabilistic Faster-RCNN to unify classification and localization adaptations into one framework. The ``source only'' results of both vanilla and Probabilistic one are presented in Table \ref{Probabilistic}, showing that the improvement brought by probabilistic modeling is trivial and limited. For a fair comparison, we have re-implemented two state-of-the-art approaches with Probabilistic Faster-RCNN in Table 1-4.
 
\begin{figure}[t]
        \centering
        \vspace{-1.0em}
        \includegraphics[width=0.45\textwidth]{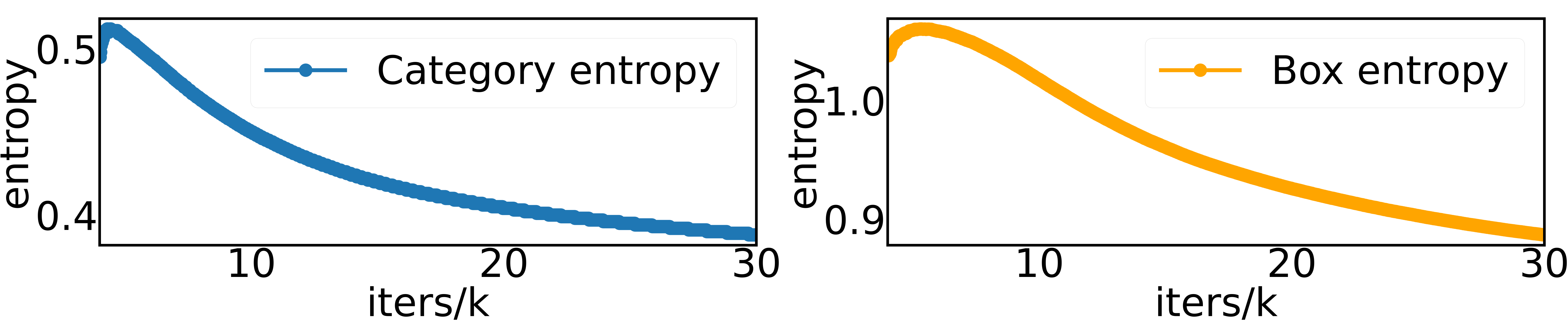}
        \vskip -0.12in
        \caption{The mean entropy of the pseudo boxes. Box entropy denotes the averaged location entropy across four coordinates.}
        \label{fig_entropy}
        \vspace{-1.0em}
\end{figure}

\textbf{The entropy of pseudo labels.}
Fig.\ref{fig_entropy} visualizes the mean entropy of the pseudo boxes on Foggy Cityscapes dataset as the training goes on in \emph{Mutual Learning} stage. We can observe that both the category and box entropy show a similar trend that they increase in a small step initially and then keep decreasing stably. It indicates that the categories of the pseudo boxes on the target domain are getting more confident, and the locations of the pseudo boxes are getting more accurate during cross-domain self-training.

\begin{figure}[t]
    \centering
    \includegraphics[width=0.45\textwidth]{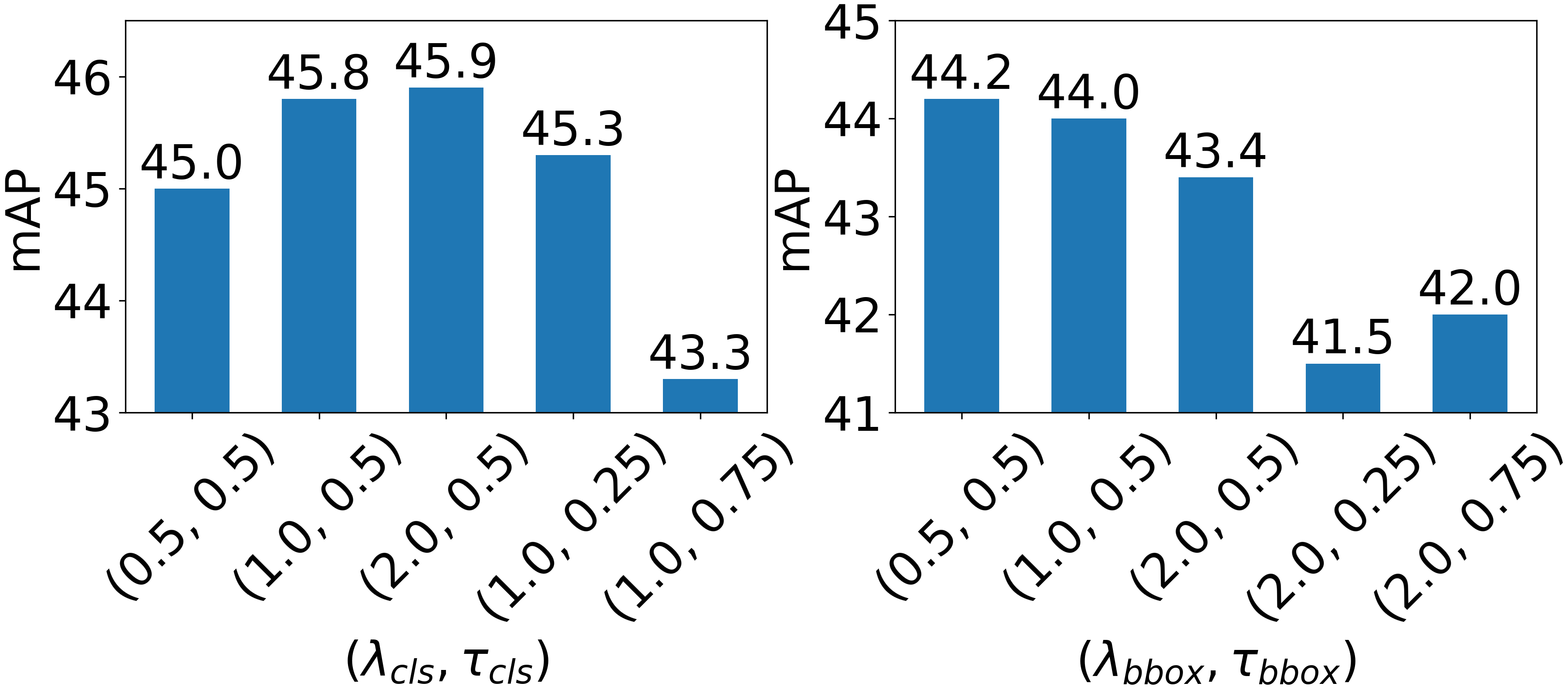}
    \vskip -0.12in
    \caption{Ablation studies of C2F on hyper parameter $\lambda$ in EFL and $\tau$ in sharpening functions.}
    \label{hyper}
    \vspace{-1.3em}
\end{figure}

\textbf{Analysis of hyper-parameters.}
Fig.\ref{hyper} provides the ablation studies of several hyper-parameters, including $\lambda$ in EFL, temperature parameters $\tau_{cls}$ and $\tau_{bbox}$. Experiments with only classification (w/o $\mathcal{L}_{T-bbox}^{RPN}+\mathcal{L}_{T-bbox}^{ROI}$, the left sub-figure of Fig.\ref{hyper}) or localization  (w/o $\mathcal{L}_{T-cls}^{RPN}+\mathcal{L}_{T-cls}^{ROI}$, the right sub-figure of Fig.\ref{hyper})  adaptation under different hyper-parameters combinations are conducted. The results show that even adapting classification or localization alone can still significantly boost performance against the ``source only''. For $\lambda$ in EFL, the adaptation performance goes up as $\lambda$ increases in classification branches while it is opposite in localization branches. For $\tau_{cls}$ and $\tau_{bbox}$, a too large or too small temperature will hurt the performance. In our paper, we set them without careful tuning (all are set to 0.5).

Furthermore, to provide a more systematic analysis about the robustness of PT w.r.t the temperature, we re-conduct the ablation study of ($\tau_{cls}$, $\tau_{bbox}$) in K2C adaptation setting, as well as a comparison with threshold-based setting. As shown in Table~\ref{comp_thre}, PT is actually more superior (see ``mean'') and robust (see ``std'') than the threshold-based method.

\textbf{Qualitative visualization.}
 As shown in Fig.\ref{fig_visualization}, we present qualitative results of C2F and K2C to demonstrate the improvement brought by PT. The visualizations show that PT can significantly ease the intra-domain gap, i.e. reducing the false-positives and increasing the true-positives. Consequently, the performance is improved by a large margin. 
See \textbf{Appendix} \ref{more_experimental_results} for more experimental results.

\vspace{-0.7em}
\section{Conclusions}
In this paper, we propose a simple yet effective framework, Probabilistic Teacher, to study the utilization of uncertainty during cross-domain self-training. Equipped with the novel Entropy Focal Loss, this framework can achieve new state-of-the-art results on multiple source-based / free UDA-OD benchmarks. We look forward that our method may bring inspirations to other weakly-supervised object detection tasks, such as noisy-label-supervised object detection.
\vspace{-0.7em}

\section*{Acknowledgement}
\vspace{-0.7em}
The work was sponsored by National Natural Science Foundation of China (U20B2066, 62106220, 62127803), Hikvision Open Fund (CCF-HIKVISION OF 20210002),
NUS Faculty Research Committee Grant (WBS: A-0009440-00-00),
and NUS Advanced Research and Technology Innovation Centre (Project Number: ECT-RP2).

\nocite{langley00}

\bibliography{ref}
\bibliographystyle{icml2022}

%%%%%%%%%%%%%%%%%%%%%%%%%%%%%%%%%%%%%%%%%%%%%%%%%%%%%%%%%%%%%%%%%%%%%%%%%%%%%%%
%%%%%%%%%%%%%%%%%%%%%%%%%%%%%%%%%%%%%%%%%%%%%%%%%%%%%%%%%%%%%%%%%%%%%%%%%%%%%%%
% APPENDIX
%%%%%%%%%%%%%%%%%%%%%%%%%%%%%%%%%%%%%%%%%%%%%%%%%%%%%%%%%%%%%%%%%%%%%%%%%%%%%%%
%%%%%%%%%%%%%%%%%%%%%%%%%%%%%%%%%%%%%%%%%%%%%%%%%%%%%%%%%%%%%%%%%%%%%%%%%%%%%%%
\newpage
\appendix
\onecolumn

\section{More Experimental Results}
\label{more_experimental_results}

\begin{figure*}[thbp]
\centering
\subfigure[Source only]
{
    \begin{minipage}[b]{.31\linewidth}
        \centering
        \includegraphics[scale=0.1]{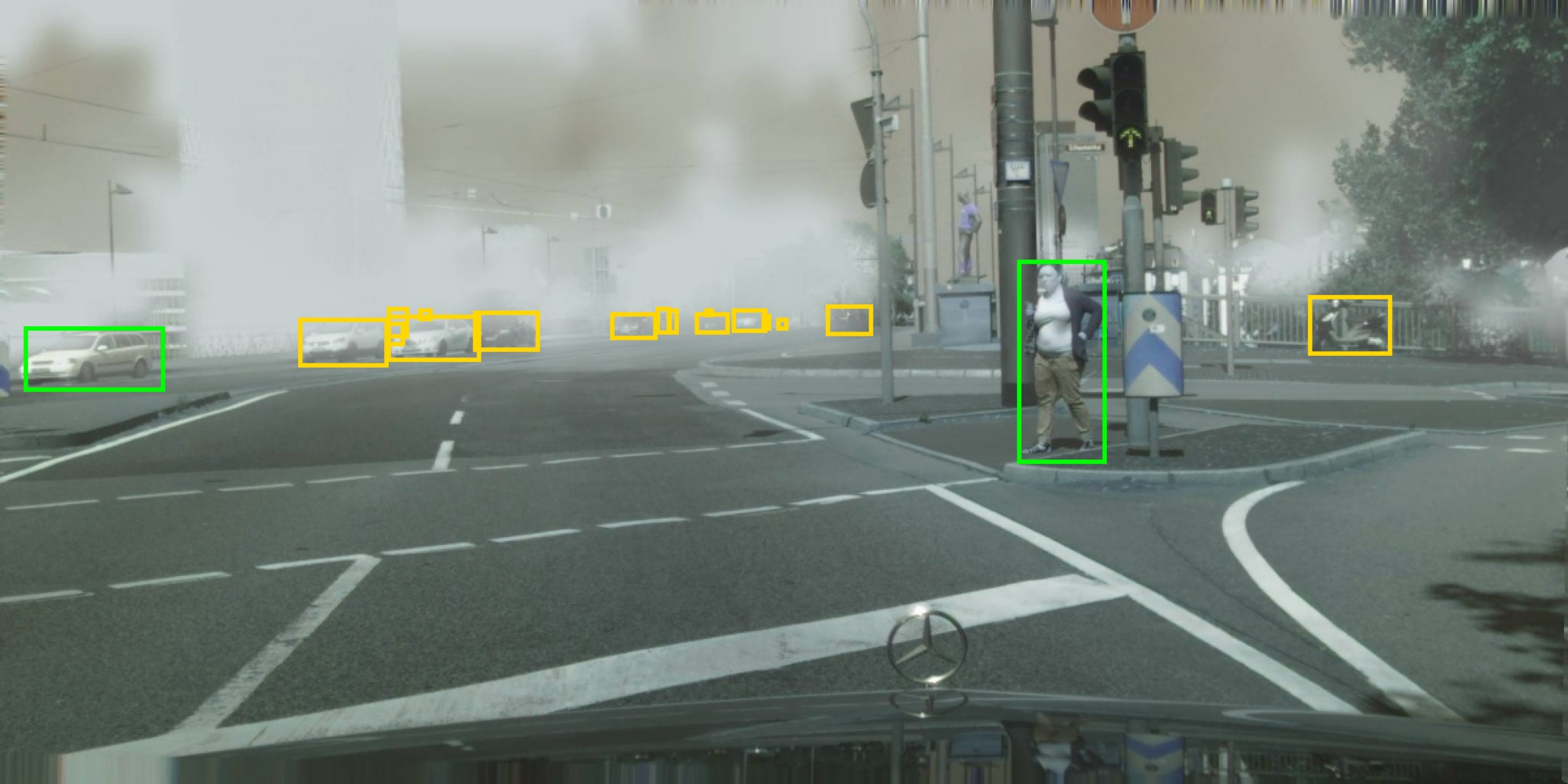} \\
        \includegraphics[scale=0.1]{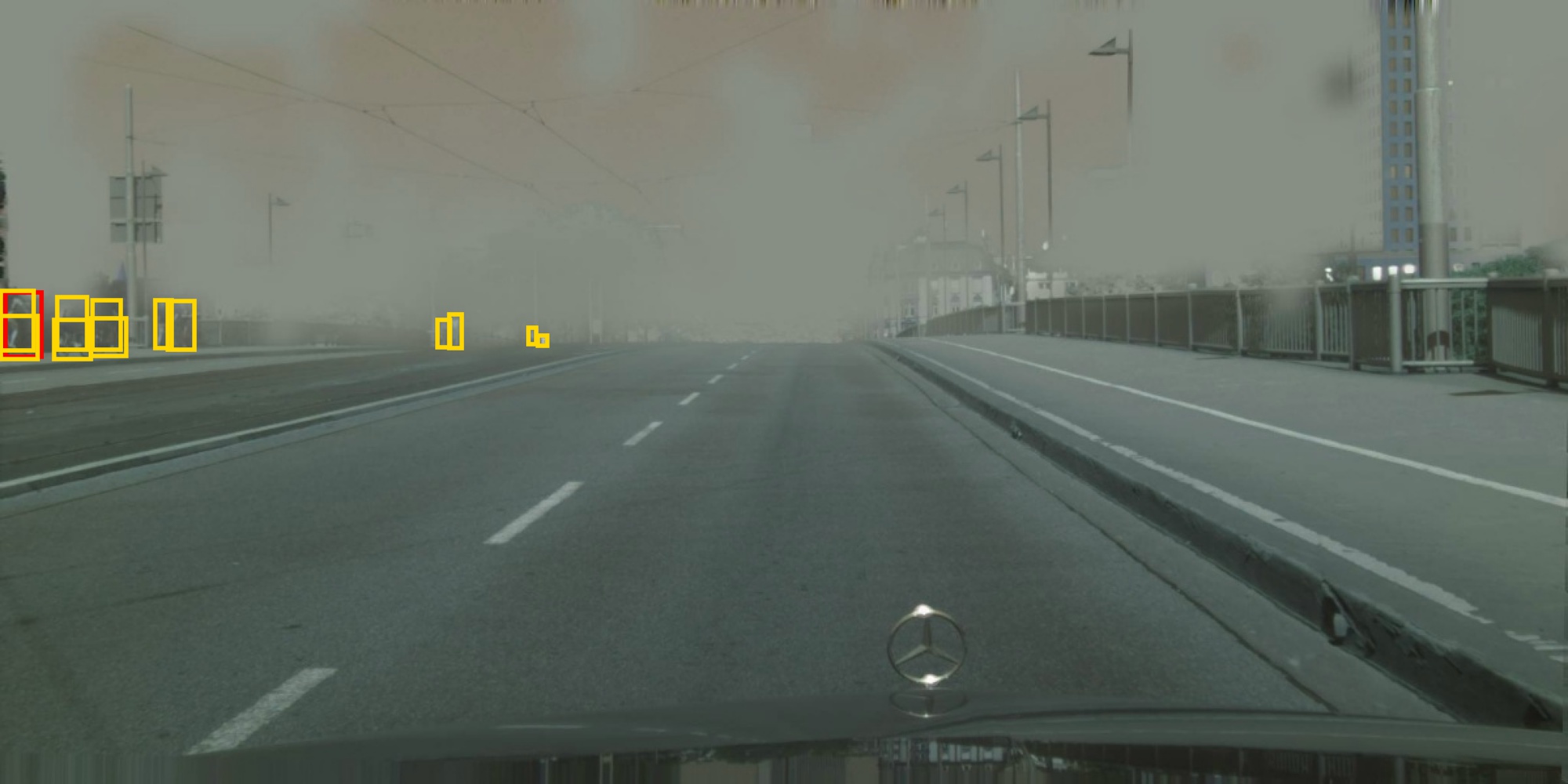} \\
        \includegraphics[scale=0.1]{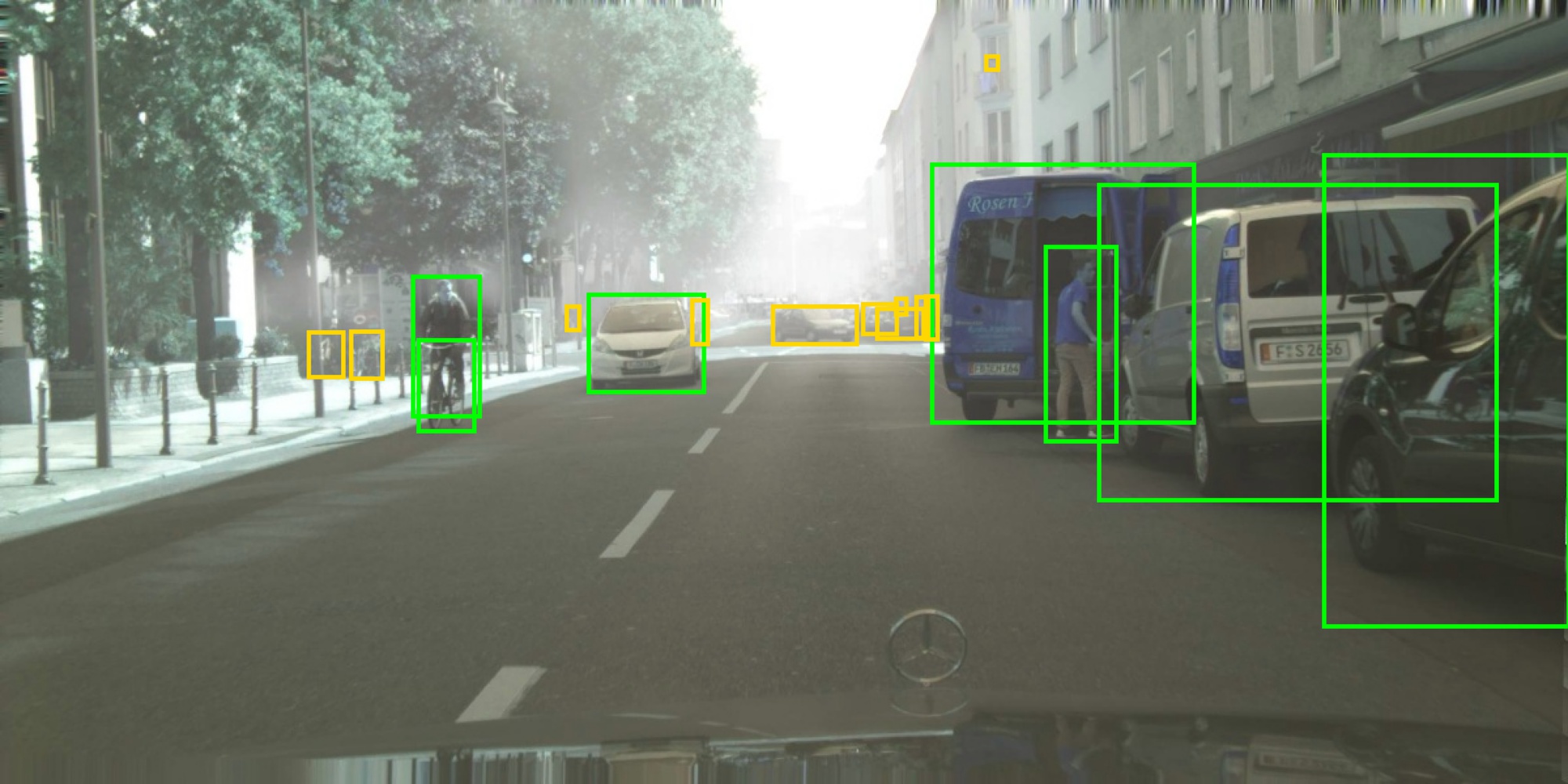} \\
    \end{minipage}
}
\centering
\subfigure[SW]
{
     \begin{minipage}[b]{.31\linewidth}
        \centering
        \includegraphics[scale=0.1]{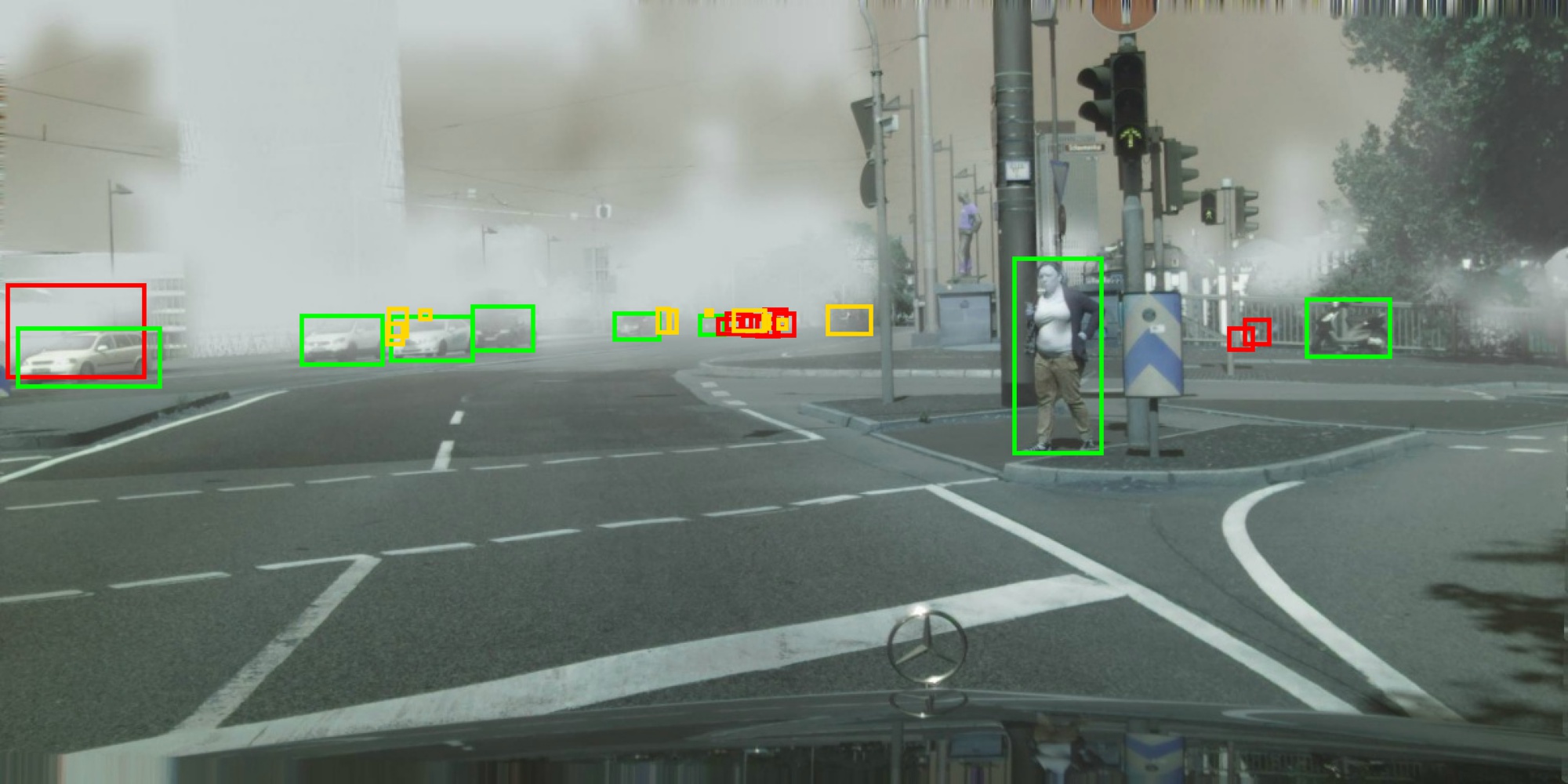} \\
        \includegraphics[scale=0.1]{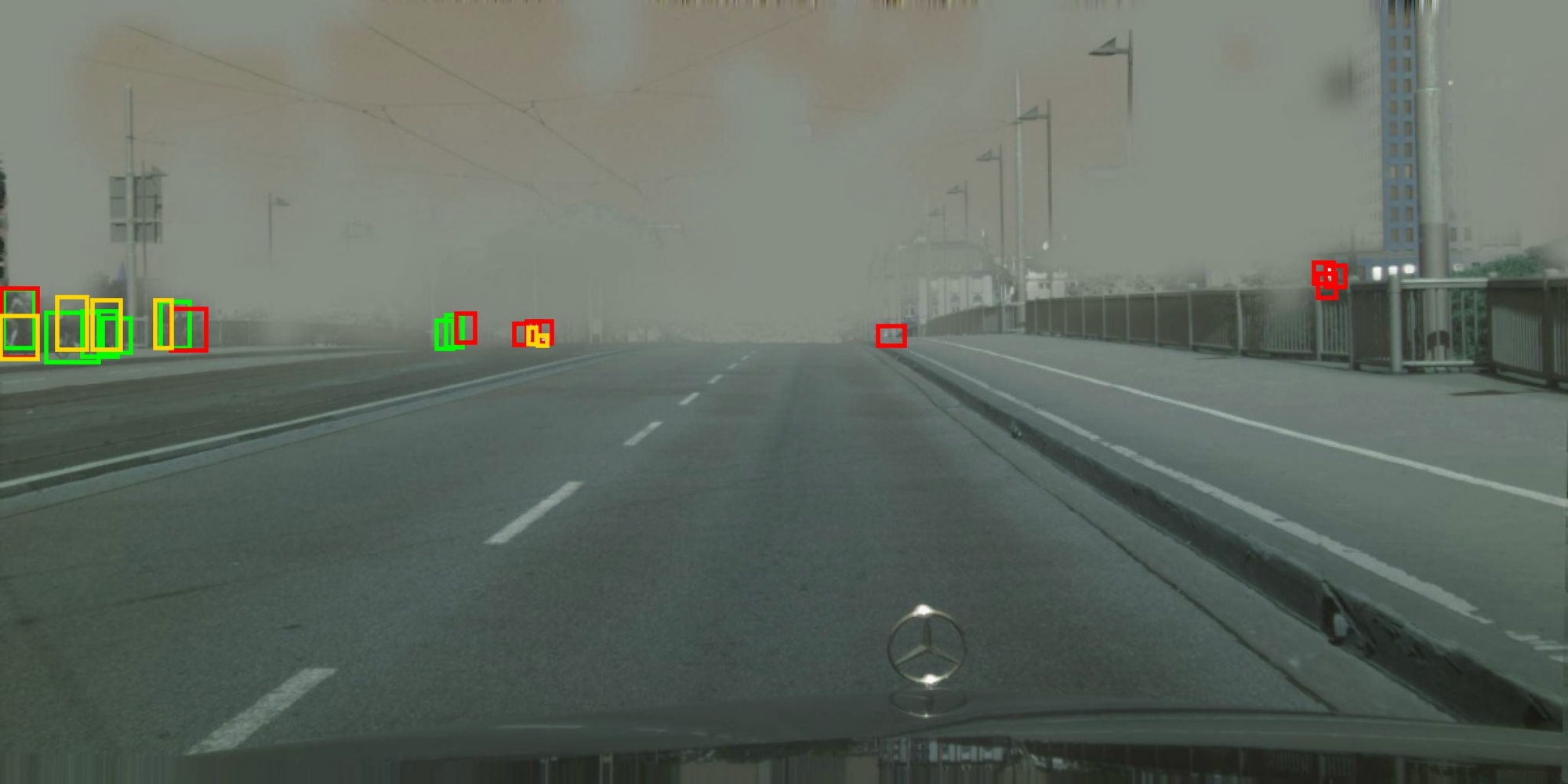} \\
        \includegraphics[scale=0.1]{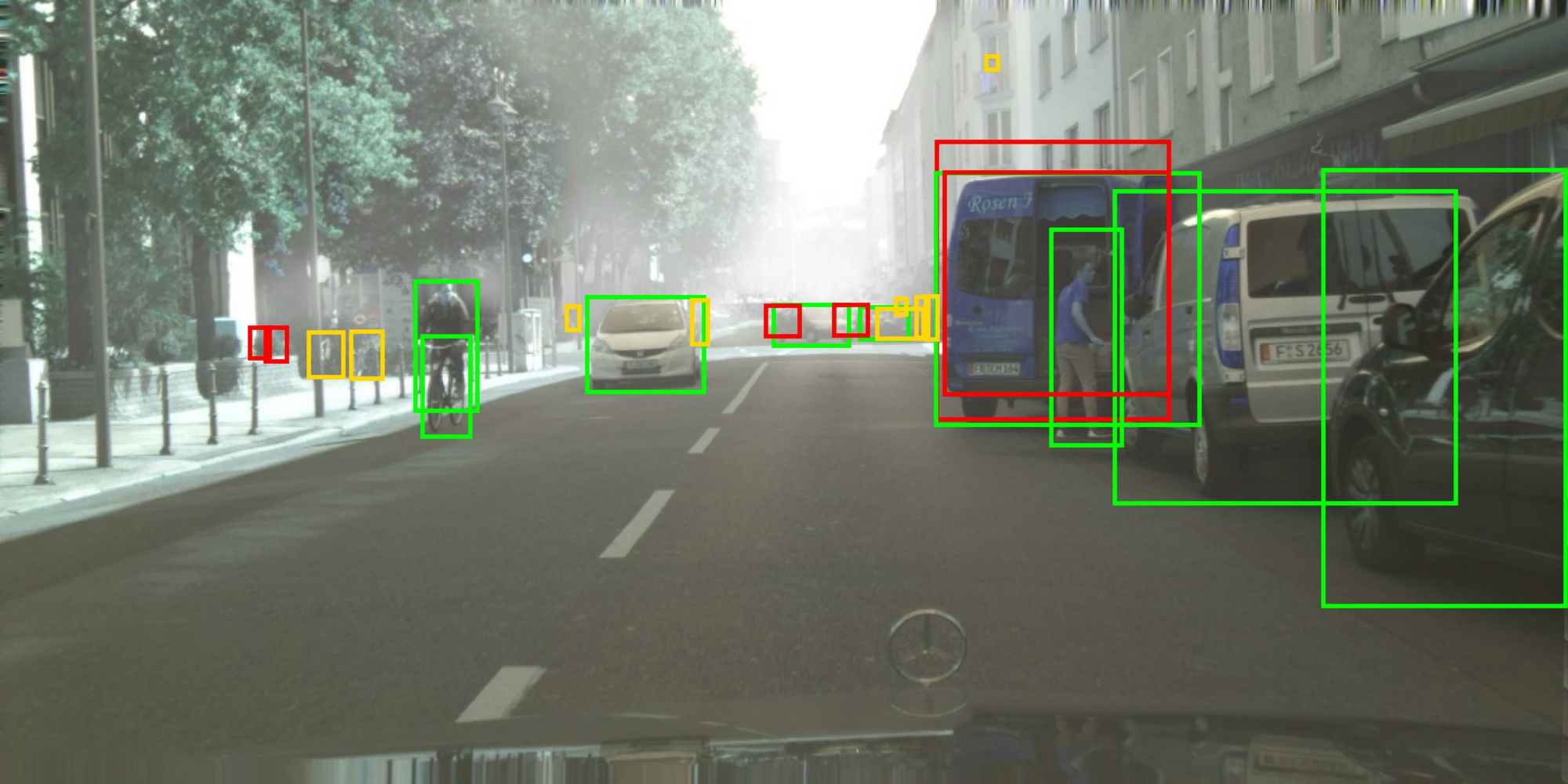} \\
    \end{minipage}
}
\centering
\subfigure[PT]
{
    \begin{minipage}[b]{.31\linewidth}
        \centering
        \includegraphics[scale=0.1]{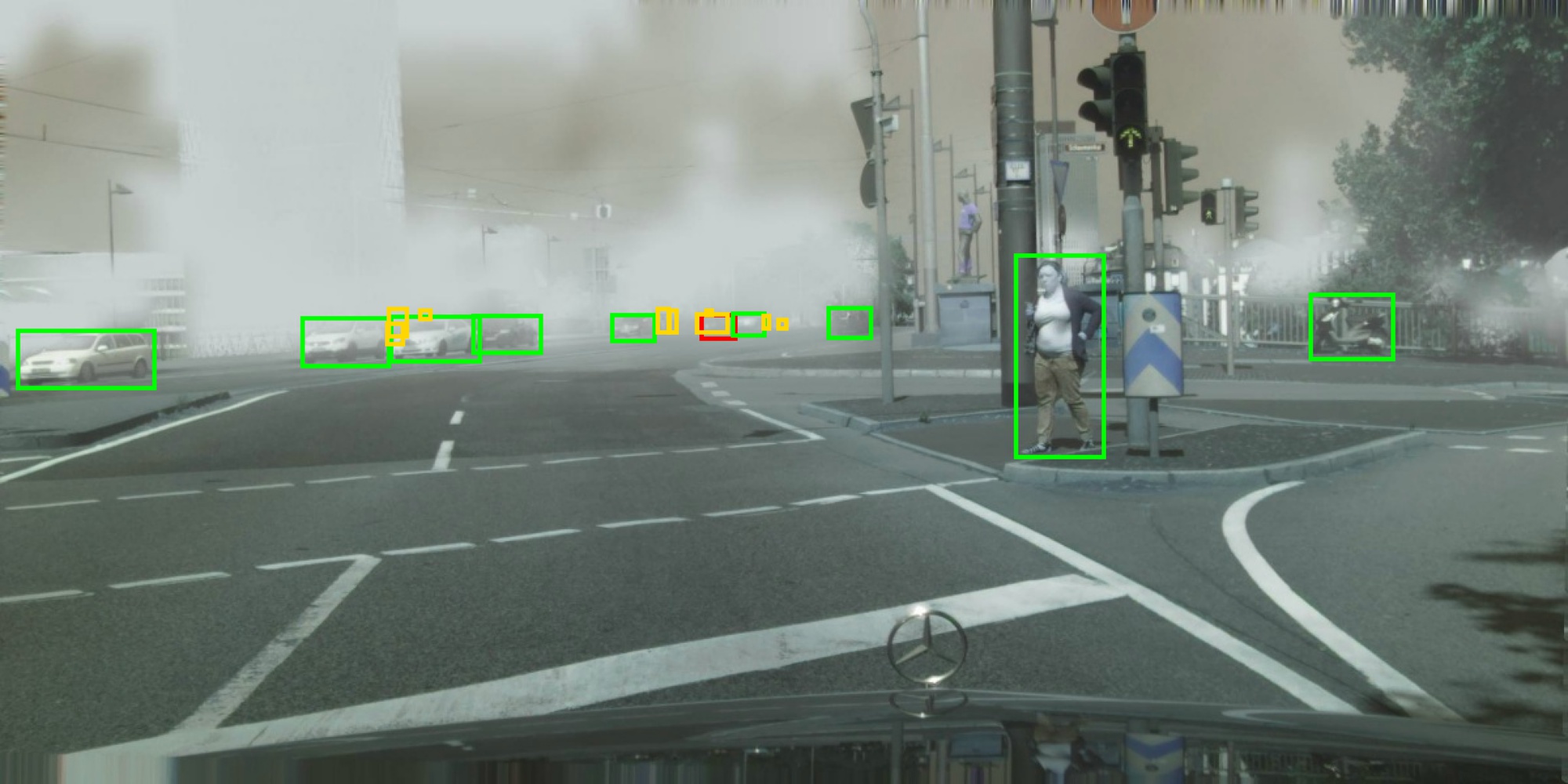} \\
        \includegraphics[scale=0.1]{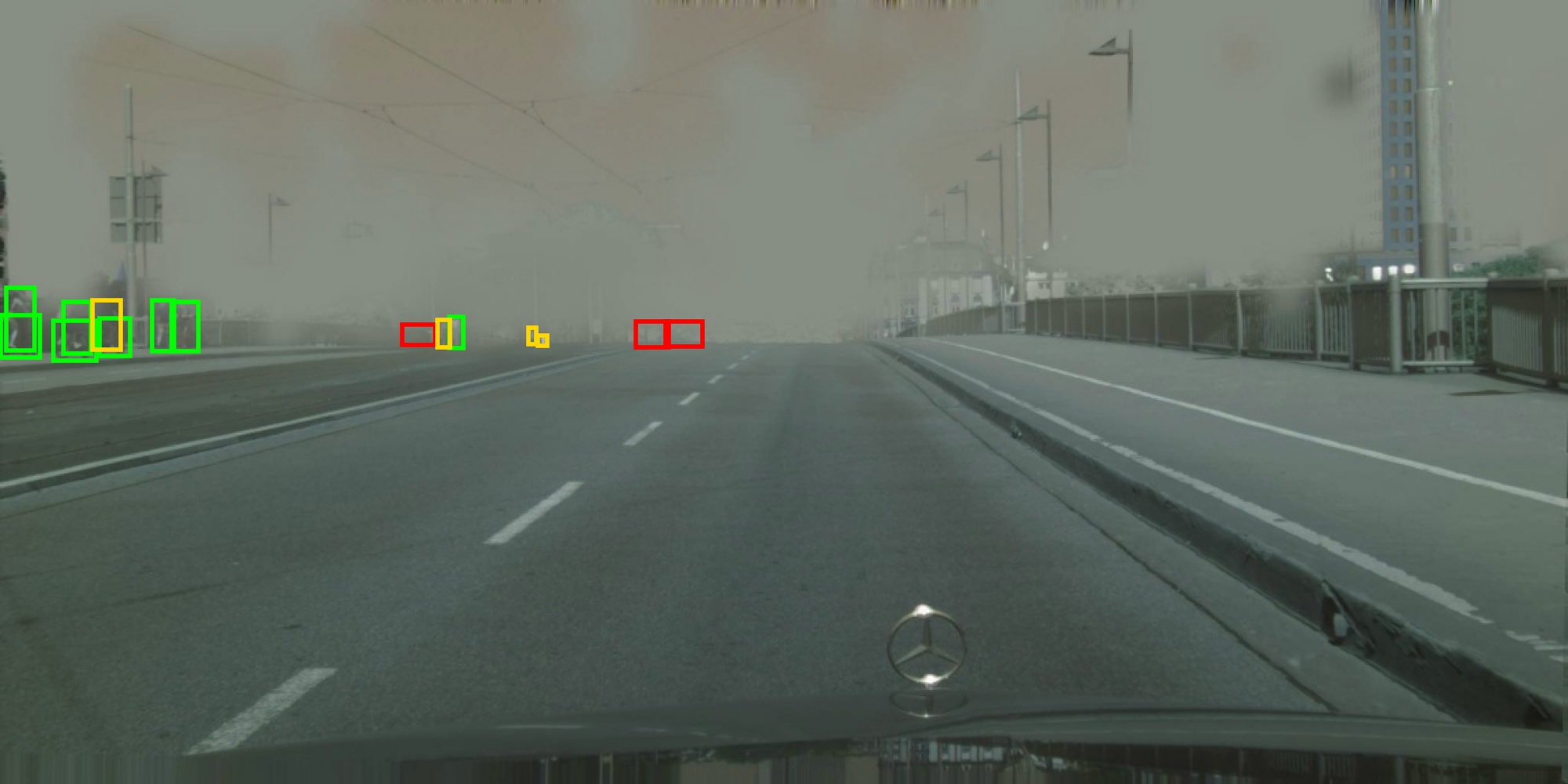} \\
        \includegraphics[scale=0.1]{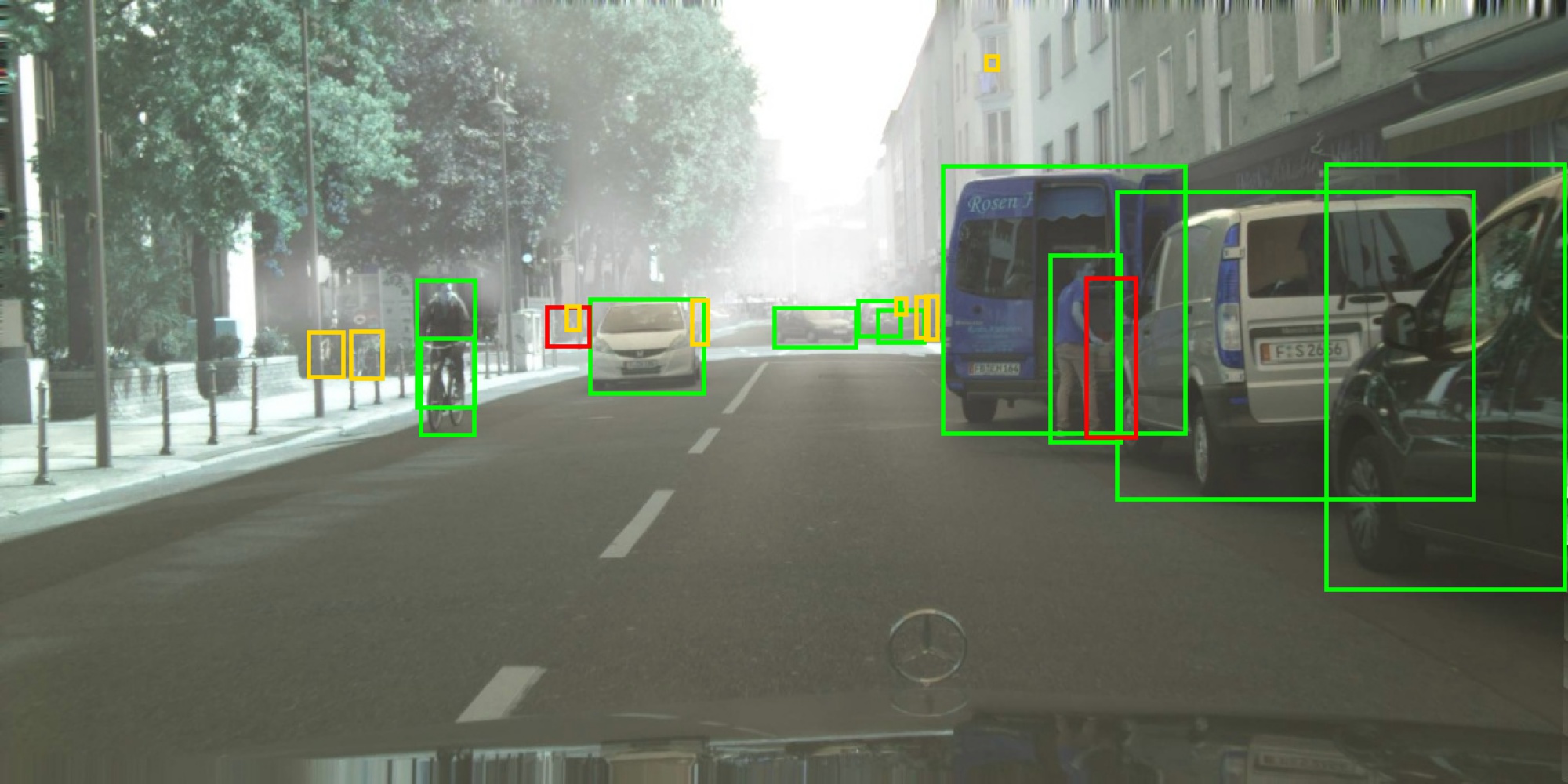} \\
    \end{minipage}
}
\caption{More results on the target domain of C2F. \textbf{\textcolor{ForestGreen}{Green}}, \textbf{\textcolor{red}{red}} and \textbf{\textcolor{Dandelion}{gold}} boxes denote true positives, false positives and false negatives, respectively.}
\label{fig_more_vis_c2f}
\end{figure*}

\begin{figure*}[h]
\centering
\subfigure[Source only]
{
    \begin{minipage}[b]{.31\linewidth}
        \centering
        \includegraphics[scale=0.1]{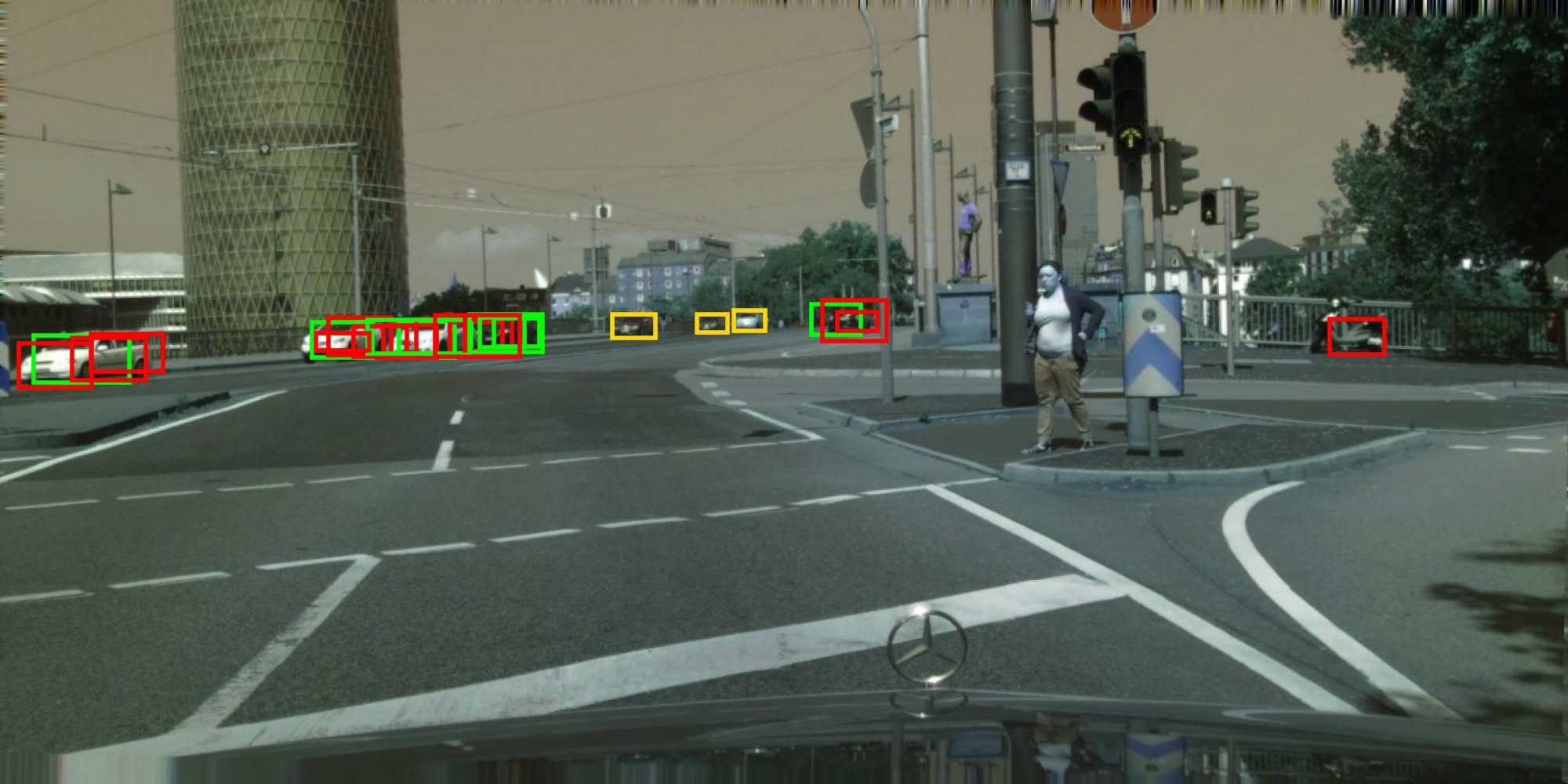} \\
        \includegraphics[scale=0.1]{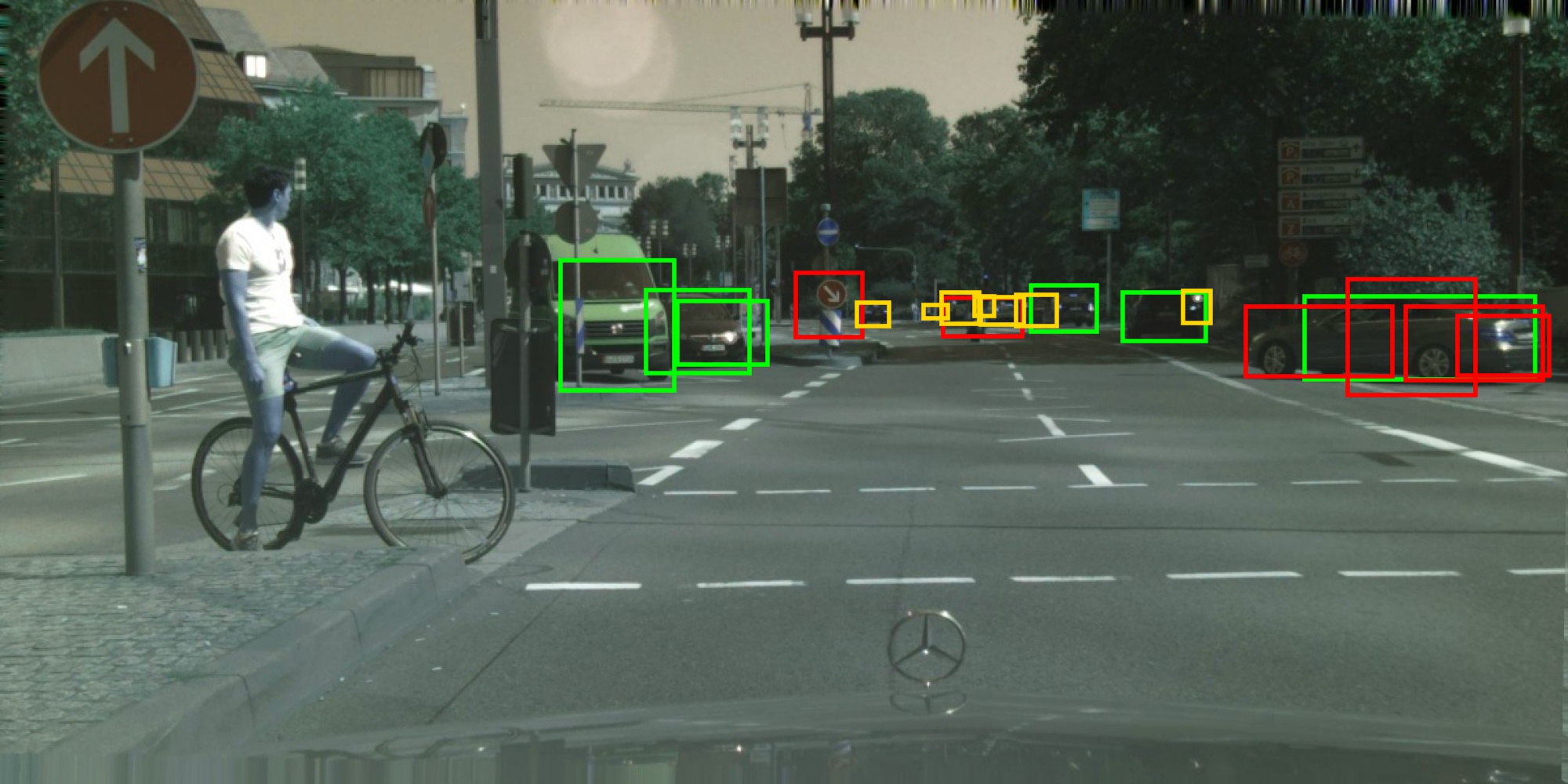} \\
        \includegraphics[scale=0.1]{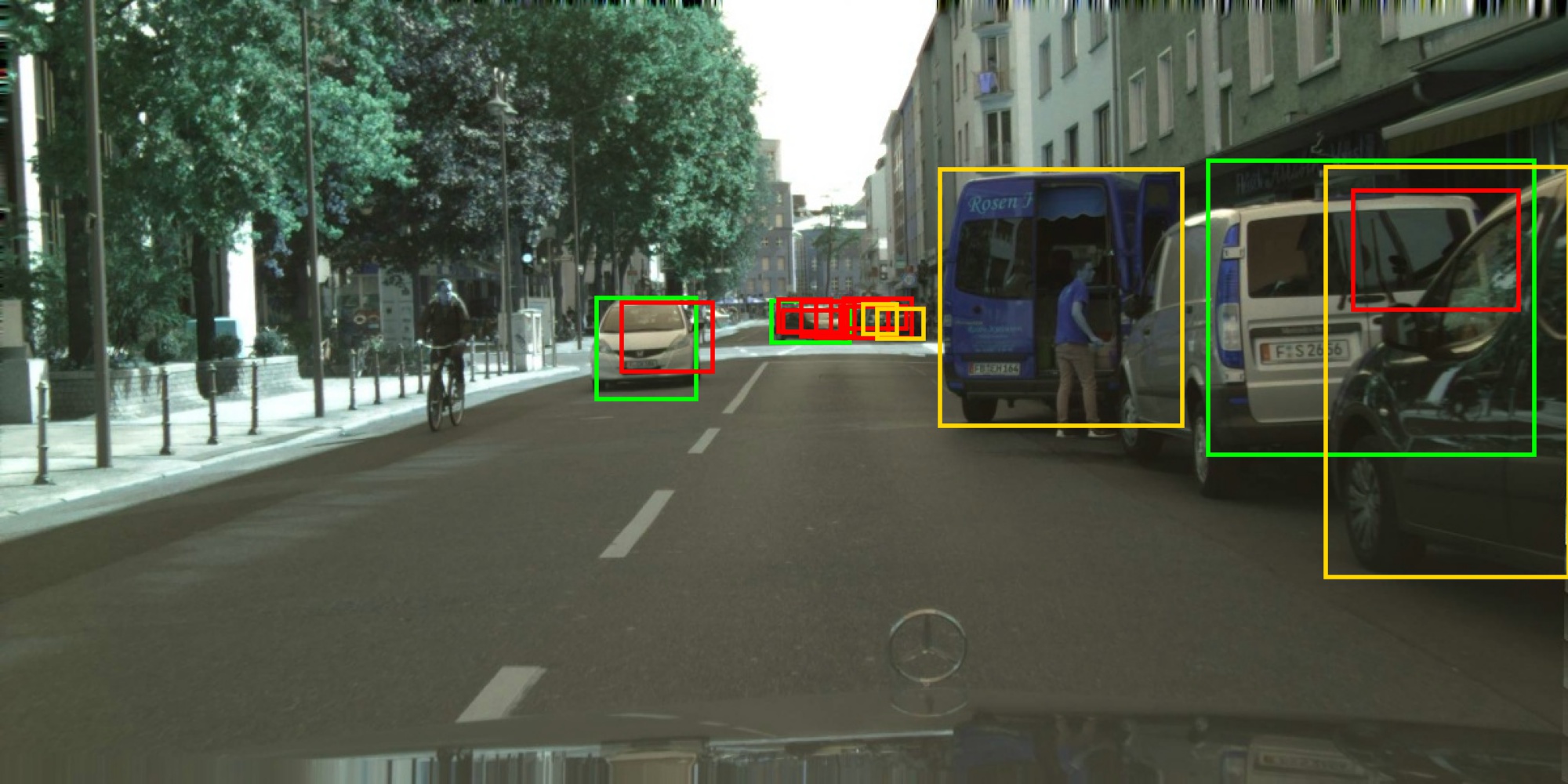} \\
    \end{minipage}
}
\centering
\subfigure[SW]
{
     \begin{minipage}[b]{.31\linewidth}
        \centering
        \includegraphics[scale=0.1]{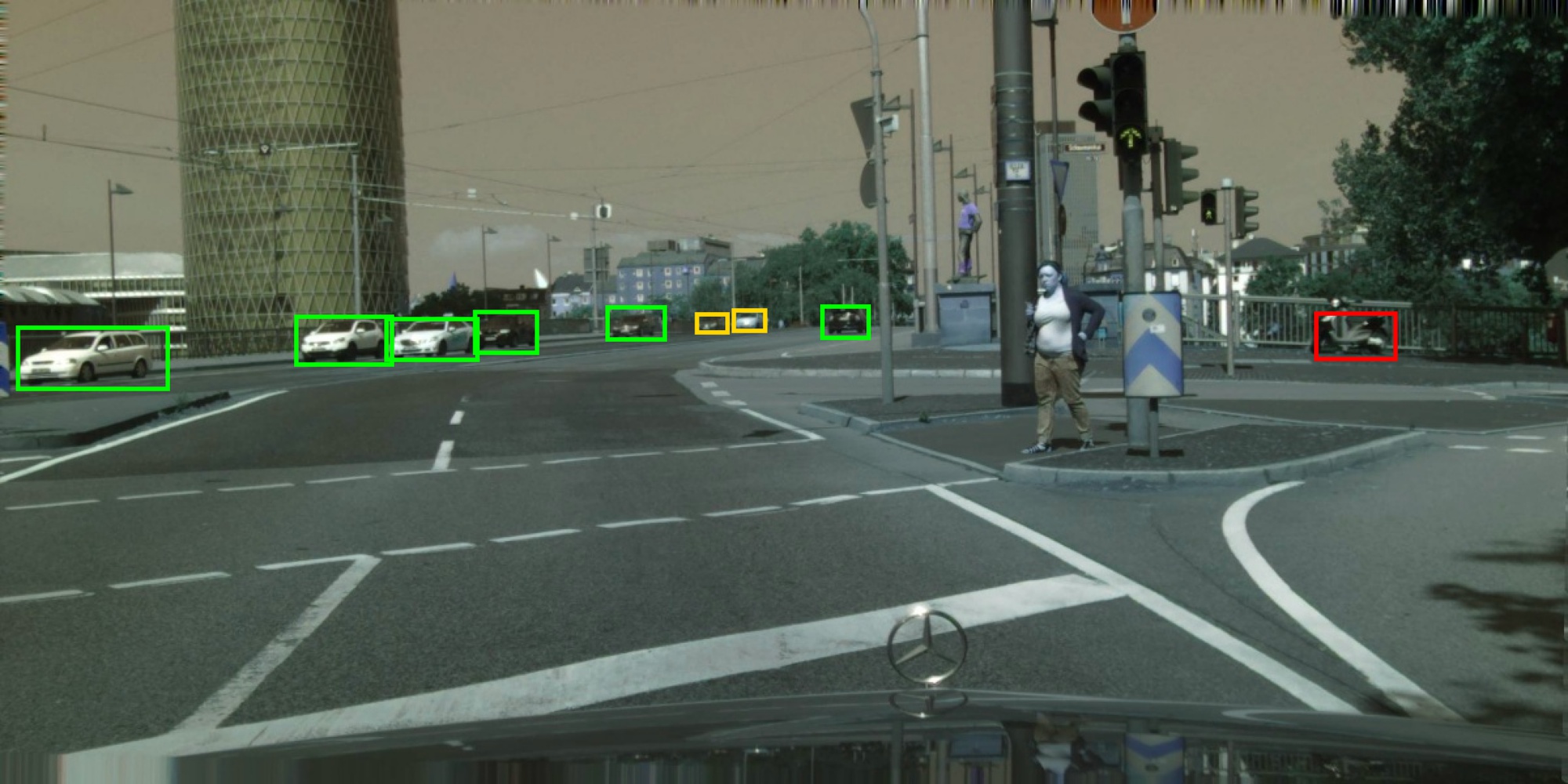} \\
        \includegraphics[scale=0.1]{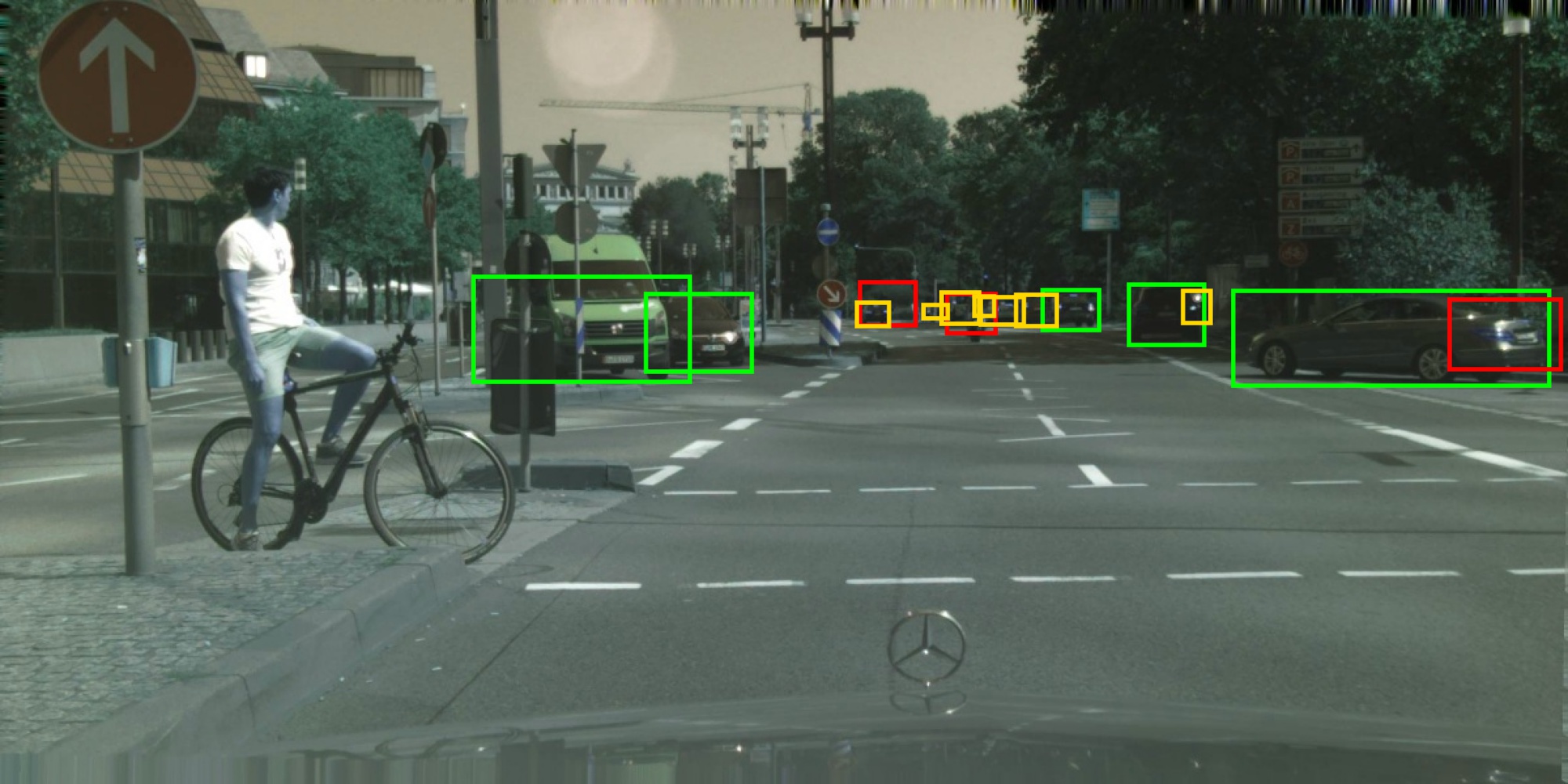} \\
        \includegraphics[scale=0.1]{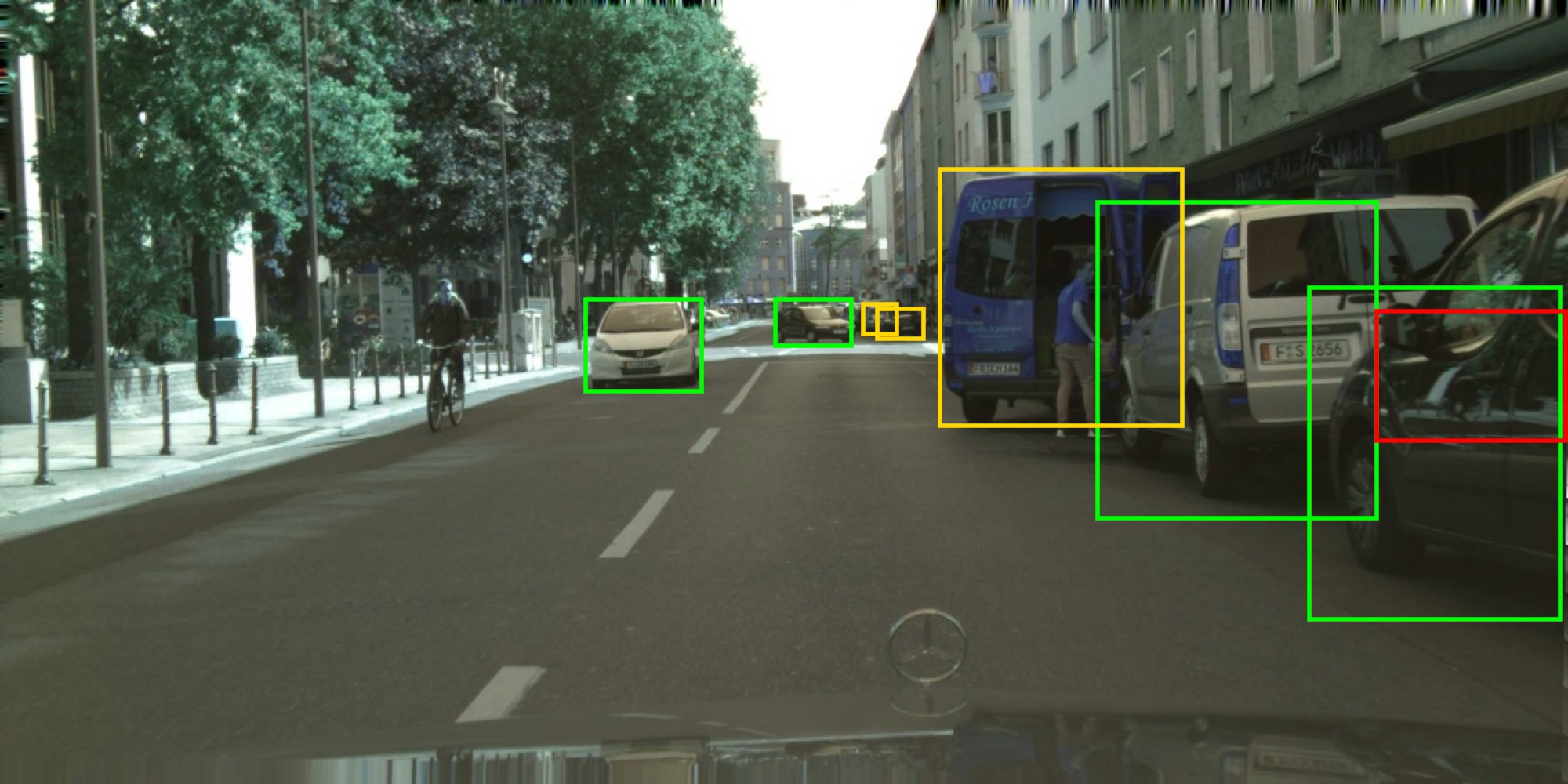} \\
    \end{minipage}
}
\centering
\subfigure[PT]
{
    \begin{minipage}[b]{.31\linewidth}
        \centering
        \includegraphics[scale=0.1]{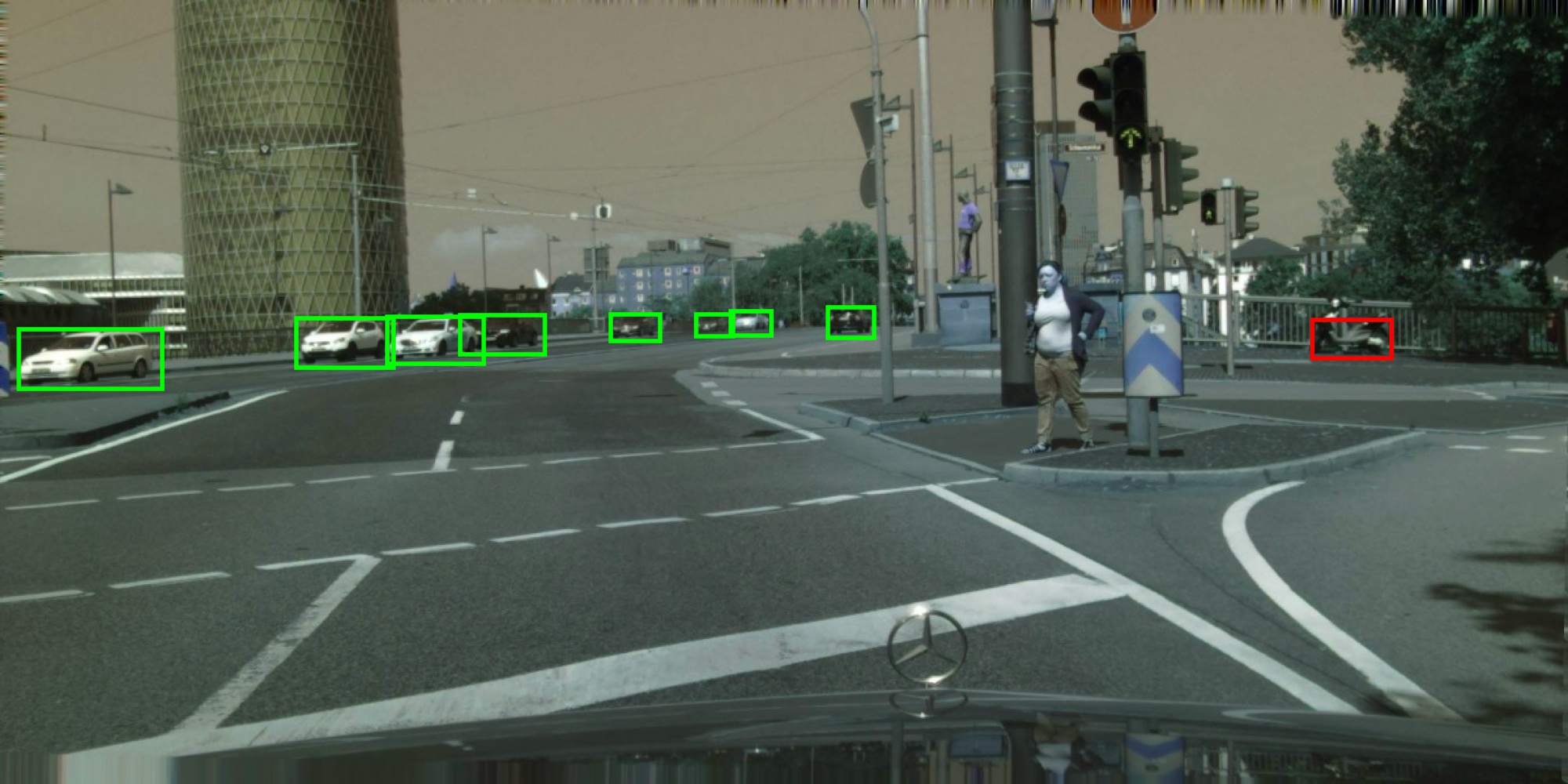} \\
        \includegraphics[scale=0.1]{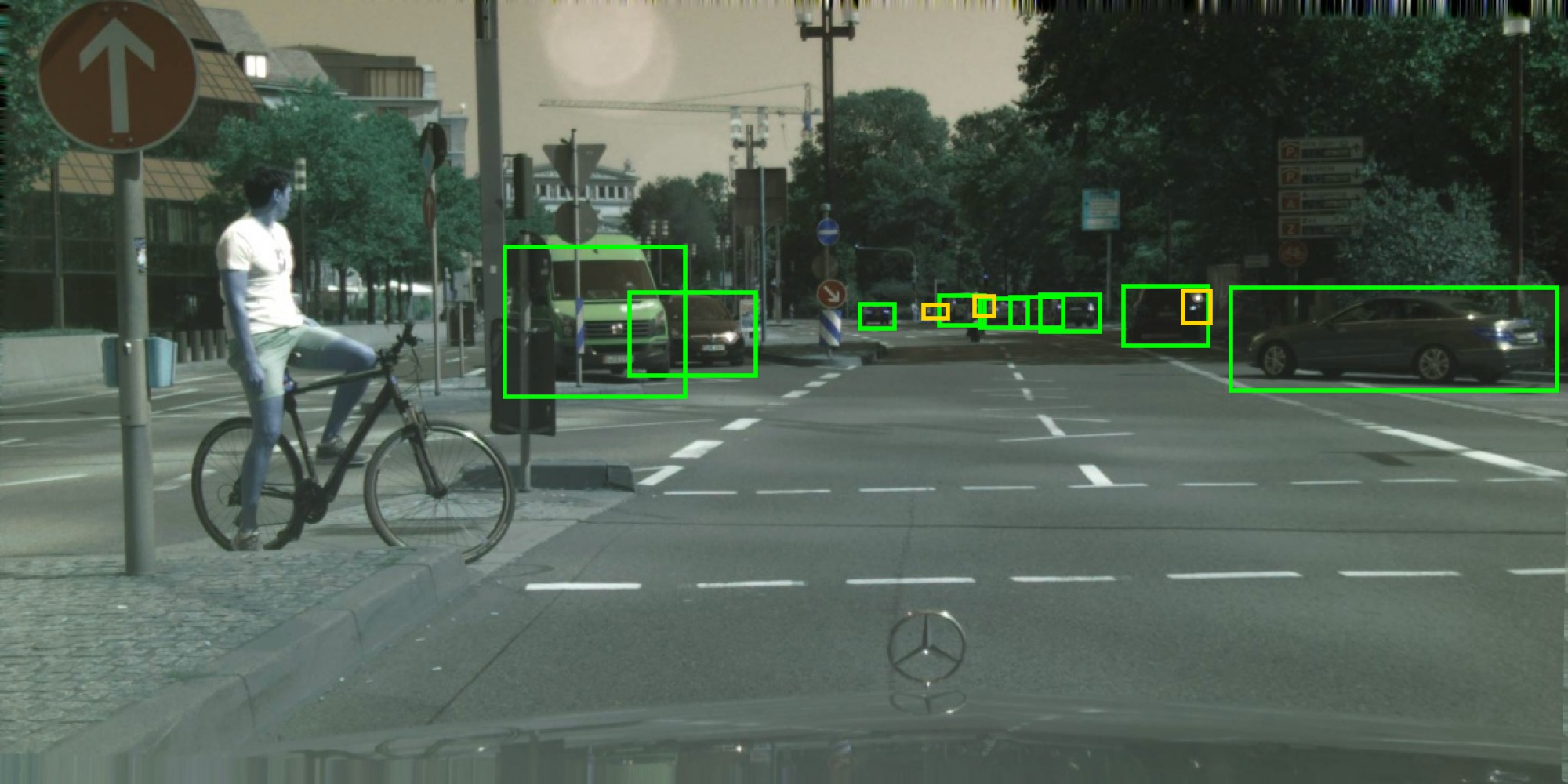} \\
        \includegraphics[scale=0.1]{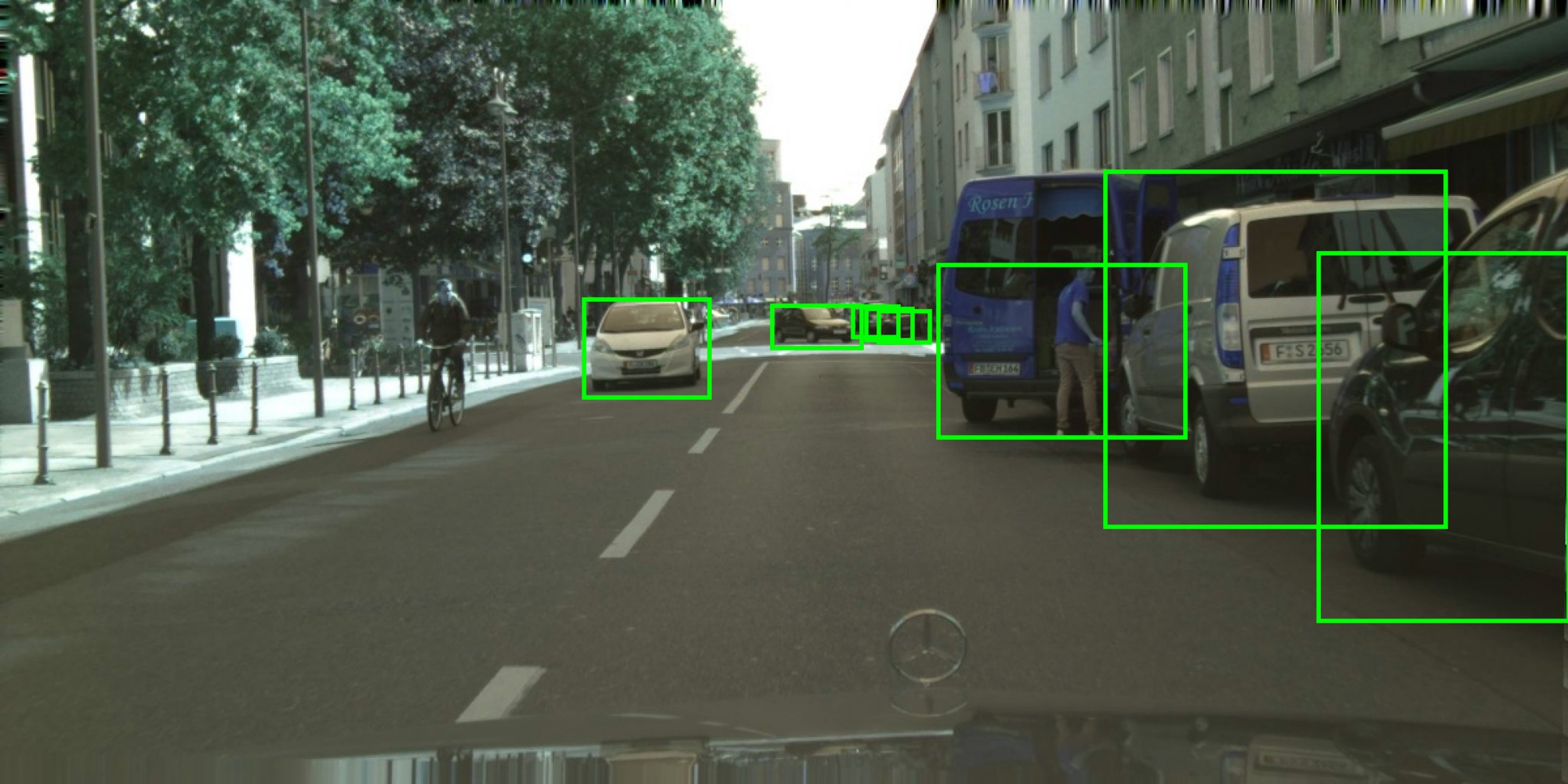} \\
    \end{minipage}
}
\caption{More results on the target domain of K2C. \textbf{\textcolor{ForestGreen}{Green}}, \textbf{\textcolor{red}{red}} and \textbf{\textcolor{Dandelion}{gold}} boxes denote true positives, false positives and false negatives, respectively.}
\label{fig_more_vis_k2c}
\end{figure*}

\subsection{More Qualitative Visualization}
In Fig.\ref{fig_more_vis_c2f} and Fig.\ref{fig_more_vis_k2c}, we present more qualitative detection results of two adaptation tasks to demonstrate the improvement brought by our PT framework. 

\section{More Implementation Details}
\label{more_implementation_details}
\subsection{Anchor adaptation}
In our implementation, the initial anchor shapes simply follow the original Faster-RCNN, i.e., 3 sizes (128, 256, 512), 3 aspects ratios (0.5, 1, 2) at each output sliding window position. In this way, only additional $9\times2$ learnable parameters are required during anchor adaptation.

\subsection{Data Augmentation Details}
In this paper, we use the same set of strong augmentation in SimCLR  \cite{28} except the RandomResizedCrop. We perform random horizontal-flipping, followed by color jitter and grayscale conversion. And then Gaussian blur and solarization are applied randomly to the images. Moreover, simple random resizing is applied to the images, which are then padded to the original size. Specifically, random resizing in our implementation is 
to zoom out the images randomly within (0.5, 1) while keeping the aspect ratios unchanged. The code for augmentation above using torchvision is as follows:

\begin{lstlisting}[language={python}, frame=shadowbox，xleftmargin=2em,xrightmargin=2em, aboveskip=1em]
from torchvision import transforms
augmentation = []
augmentation.append(transforms.RandomApply([HorizontalFlipping()], p=0.5))
augmentation.append(transforms.RandomApply([transforms.ColorJitter(0.4, 0.4, 0.4, 0.1)], p=0.8))
augmentation.append(transforms.RandomGrayscale(p=0.2))
augmentation.append(transforms.RandomApply([GaussianBlur([0.1, 2.0])], p=0.5))
augmentation.append(transforms.RandomApply([Solarize(threshold=0.5)], p=0.2))
augmentation.append(transforms.RandomApply([RandomResizing(0.5, 1.0)], p=1.0))
\end{lstlisting}

\subsection{Pseudo-Code of PT}
Algorithm 1 summarizes the pseudo-code of PT. PT consists of two training stages, including \emph{Pretraining} and \emph{Mutual Learning}. In the \emph{Pretraining}  stage, we train the detector using the labeled source data to initialize the detector, and then duplicate the trained weights to both the teacher and student models. In the \emph{Mutual Learning} stage, the generated pseudo boxes on the unlabeled target data and the labeled source data are used to train the student via uncertainty-guided consistency training for both classification and localization branches. The student then transfers its learned knowledge to the teacher via EMA. In this way, both models can evolve jointly and continuously to improve performance.

\begin{algorithm}[tbh]
	\renewcommand{\algorithmicrequire}{\textbf{Input:}}
	\caption{Probabilistic Teacher}
	\label{alg1}
	\begin{algorithmic}[1]
	    \REQUIRE Source domain $D_{S}$, Target domain $D_{T}$
		\WHILE{Pretraining} 
    		\STATE Train source only model $\theta^{I}$ based on Eq.~(3)
    	\ENDWHILE
    	\STATE $\theta^{S} \leftarrow \theta^{I}$  \COMMENT{Duplicate to student model}
    	\STATE $\theta^{T} \leftarrow \theta^{I}$  \COMMENT{Duplicate to teacher model}
		\WHILE{Mutual Learning}
    		\STATE Calculate $\mathcal{L}_{S}$ based on Eq.~(3)
    		\STATE $p^{PL}, t^{PL} \leftarrow \theta^{T}(D_{T})$  \COMMENT{Pseudo labeling, weak augmentation}
    	    \STATE $p^{PL} \leftarrow \mathcal{S}_{cls}(p^{PL}, \tau_{cls})$  \COMMENT{Sharpen classification probability distributions}
    	    \STATE $t^{PL} \leftarrow \mathcal{S}_{bbox}(t^{PL}, \tau_{bbox})$  \COMMENT{Sharpen localization probability distributions}
    	    \STATE $p, t \leftarrow \theta^{S}(D_{T})$  \COMMENT{Feed forward, strong augmentation}
    	    \STATE Calculate EFL $\mathcal{L}_{T-cls}^{ROI}$ based on Eq.~(6) and Eq.~(11) \COMMENT{Classification adaptation}
    	    \STATE $p^{PL} \leftarrow \mathcal{M}(p^{PL})$ \COMMENT{Merging operation to sum up all foreground probabilities}
    	    \STATE Calculate EFL $\mathcal{L}_{T-cls}^{RPN}$ based on Eq.~(6) and Eq.~(11) \COMMENT{Classification adaptation}
    	    \STATE Calculate EFL $\mathcal{L}_{T-bbox}^{ROI}$ based on Eq.~(7) and Eq.~(11) \COMMENT{Regression adaptation}
    	    \STATE Calculate EFL $\mathcal{L}_{T-bbox}^{RPN}$ based on Eq.~(7) and Eq.~(11) \COMMENT{Regression adaptation}
    	    \STATE $\mathcal{L}_{T}=\mathcal{L}_{T-cls}^{RPN}+\mathcal{L}_{T-cls}^{ROI}+\mathcal{L}_{T-box}^{RPN}+\mathcal{L}_{T-box}^{ROI}$
    	    \STATE $\mathcal{L}_{total}=\mathcal{L}_{S}+\lambda_{T} \mathcal{L}_{T}$
    	    \STATE Train the anchor shapes via minimizing $\mathcal{L}_{T}$  \COMMENT{Anchor adaptation}
    	    \STATE Train student model $\theta^{S}$ via minimizing $\mathcal{L}_{total}$
    	    \STATE Update teacher model via EMA
    	\ENDWHILE
	\end{algorithmic}  
\end{algorithm}

\section{Mathematical Proofs}
In this section, we provide the rigorous proofs mentioned in the main body of the paper.

\subsection{The cross-entropy between a Dirac delta distribution and a univariate Gaussian distribution}
\label{appendix_Dirac_Gaussian}
Given $ p(x) = \delta(x-a)$  and $ q(x) =  N(x; \mu, \sigma^{2})$,
the cross-entropy between $ p(x) $ and  $ q(x) $, $H(p, q)$, can be written as:
$$
\begin{aligned}
H(p, q)&=-\int p(x) \log q(x) d x \\
& =-\int \delta(x-a) \log (N(x; \mu, \sigma^{2})) d x
\end{aligned}
$$
Expanding the $\log$ term,
$$
\begin{aligned}
H(p, q)&=-\int \delta(x-a) \log (N(x; \mu, \sigma^{2})) d x \\
& = \frac{1}{2} \int \delta(x-a) \left(\log (2 \pi)+2 \log \sigma+\left(\frac{x-\mu}{\sigma}\right)^{2}\right) d x
\end{aligned}
$$
Because,
$$
\begin{aligned}
\delta(x - a) &=0,(x \neq a) \\
\int\delta(x-a) d x &=1
\end{aligned}
$$
The cross-entropy between $ p(x) $ and  $ q(x) $, $H(p, q)$, can be simplified as:
$$
\begin{aligned}
& H(p, q) = \frac{1}{2}\left(\log (2 \pi)+2 \log \sigma+\left(\frac{a-\mu}{\sigma}\right)^{2}\right)
\end{aligned}
$$

\subsection{The entropy of a univariate Gaussian distribution}
\label{appendix_entropy_Gaussian}
Let $x$ be a univariate Gaussian distributed random variable:
$$x \sim \textit{N}\left(x; \mu, \sigma^{2}\right)$$
The differential entropy of $x$, $H(x)$, can be written as:
$$
\begin{aligned}
H(x)&=-\int p(x) \log p(x) \mathbf{d} x \\
&=\frac{1}{2} \log \left(2 \pi \sigma^{2}\right)+\frac{1}{2 \sigma^{2}} \mathrm{E}\left[(x-\mu)^{2}\right]
\end{aligned}
$$
The expectation of $(x-\mu)^{2}$, ${E}[($x$-\mu)^{2}]$, is equal to the variance:
$$
E[(x-\mu)^{2}] = \sigma^{2}
$$
Substituting this back in the earlier expression gives us the result,
$$
H(x) {=} \frac{1}{2} \log \left(2 \pi \sigma^{2}\right)+\frac{1}{2}
$$
The entropy of a univariate Gaussian distribution is only the function of its variance. The maximal value in our paper is $\frac{1}{2} \log \left(2 \pi \right)+\frac{1}{2}$ since $\sigma$ is processed as a value between zero and one with a sigmoid function.

\subsection{The cross-entropy between two univariate Gaussian distributions}
\label{appendix_two_Gaussian}
Given $ p(x) = \textit{N}\left(x; \mu_{1}, \sigma_{1}^{2}\right)$  and $ q(x) = \textit{N}\left(x; \mu_{2}, \sigma_{2}^{2}\right)$ , the cross-entropy between $ p(x) $ and  $ q(x) $, $H(p, q)$, can be written as,
$$
\begin{aligned}
H(p, q)&=-\int p(x) \log q(x) d x \\
& =-\int \textit{N}\left(x; \mu_{1}, \sigma_{1}^{2}\right) \log (\textit{N}\left(x; \mu_{2}, \sigma_{2}^{2}\right)) d x
\end{aligned}
$$
Expanding the $\log$ term,
$$
\begin{aligned}
H(p, q)&=-\int \textit{N}\left(x; \mu_{1}, \sigma_{1}^{2}\right) \log (\textit{N}\left(x; \mu_{2}, \sigma_{2}^{2}\right)) d x \\
& = \frac{1}{2} \int \textit{N}\left(x; \mu_{1}, \sigma_{1}^{2}\right) \left(\log (2 \pi)+2 \log \sigma_{2}+\left(\frac{x-\mu_{2}}{\sigma_{2}}\right)^{2}\right) d x
\end{aligned}
$$
Because the integral over a PDF is always 1,
$$
\int\textit{N}\left(x; \mu, \sigma^{2}\right) d x=1
$$
Moving the constant outside,
$$
H(p, q)=\frac{1}{2}\left(\log (2 \pi)+2 \log \sigma_{2}\right)+ \frac{1}{2} \int \textit{N}\left((x; \mu_{2}, \sigma_{2}^{2}\right) \left(\frac{x-\mu_{2}}{\sigma_{2}}\right)^{2} d x
$$
Now let's only consider the second term. Because,
$$
\left(\frac{x-\mu_{2}}{\sigma_{2}}\right)^{2}=\left(\frac{x-\mu_{1}}{\sigma_{1}}\right)^{2} \frac{\sigma_{1}^{2}}{\sigma_{2}^{2}}+x \frac{2\left(\mu_{1}-\mu_{2}\right)}{\sigma_{2}^{2}}+\frac{\mu_{2}^{2}-\mu_{1}^{2}}{\sigma_{2}^{2}}
$$
The second term can be expanded as,
$$
\begin{aligned}
\frac{1}{2} \int \textit{N}\left(x; \mu_{2}, \sigma_{2}^{2}\right) \left(\frac{x-\mu_{2}}{\sigma_{2}}\right)^{2} d x &=\frac{1}{2} \int \textit{N}\left(x; \mu_{2}, \sigma_{2}^{2}\right) \left(\left(\frac{x-\mu_{1}}{\sigma_{1}}\right)^{2} \frac{\sigma_{1}^{2}}{\sigma_{2}^{2}}+x \frac{2\left(\mu_{1}-\mu_{2}\right)}{\sigma_{2}^{2}}+\frac{\mu_{2}^{2}-\mu_{1}^{2}}{\sigma_{2}^{2}}\right) d x \\
= & \frac{1}{2\sigma_{2}^{2}} \int \textit{N}\left(x; \mu_{2}, \sigma_{2}^{2}\right) \left(x-\mu_{1}\right)^{2} d x + \frac{\left(\mu_{1}-\mu_{2}\right)}{\sigma_{2}^{2}}\int \textit{N}\left(x; \mu_{2}, \sigma_{2}^{2}\right) x d x \\
& + \frac{1}{2}\frac{\mu_{2}^{2}-\mu_{1}^{2}}{\sigma_{2}^{2}}\int \textit{N}\left(x; \mu_{2}, \sigma_{2}^{2}\right) d x \\
\end{aligned}
$$
Because, 1) the integral over a PDF is always 1; 2) the expectation of x, E(x), is equal to the mean; 3) the expectation of $(x-\mu)^{2}$, ${E}[($x$-\mu)^{2}]$, is equal to the variance, and these are,
$$
\begin{aligned}
\int\textit{N}\left(x; \mu, \sigma^{2}\right) d x &=1 \\
\int x N(x ; \mu,  \sigma^{2}) d x &=\mu \\
\int(x - \mu)^2 N(x ; \mu,  \sigma^{2}) d x &= \sigma^{2}
\end{aligned}
$$
Therefore,
$$
\begin{aligned}
\frac{1}{2} \int \textit{N}\left(x; \mu_{2}, \sigma_{2}^{2}\right) \left(\frac{x-\mu_{2}}{\sigma_{2}}\right)^{2} d x =\frac{1}{2}\left(\frac{\sigma_{1}^{2}}{\sigma_{2}^{2}}+\frac{2 \mu_{1}\left(\mu_{1}-\mu_{2}\right)}{\sigma_{2}^{2}}+\frac{\mu_{2}^{2}-\mu_{1}^{2}}{\sigma_{2}^{2}}\right)
\end{aligned}
$$
Substituting this back in the earlier expression gives us the result,
$$
\begin{aligned}
H(p, q)&=\frac{1}{2}\left(\log (2 \pi)+2 \log \sigma_{2}+\frac{\sigma_{1}^{2}}{\sigma_{2}^{2}}+\frac{2 \mu_{1}\left(\mu_{1}-\mu_{2}\right)}{\sigma_{2}^{2}}+\frac{\mu_{2}^{2}-\mu_{1}^{2}}{\sigma_{2}^{2}}\right) \\
& = \log \sigma_{2}+\frac{\sigma_{1}^{2} + (\mu_{1}-\mu_{2})^2}{2\sigma_{2}^{2}} + C
\end{aligned}
$$
where C is a constant.

\subsection{ Maximal value of a discrete variable's entropy}
\label{appendix_Maximal_entropy}
Let $x$ be a discrete variable,
$$x \sim p(x)$$
Its entropy, $H(x)$, can be written as,
$$
\begin{aligned}
H(x)&=- \sum p(x) \log p(x) \\
& = E[ \log \frac{1}{p(x)}]
\end{aligned}
$$
This means that the entropy of  $x$ is equal to the expectation of $\log \frac{1}{p(x)}$.
Using the Jensen inequality,
$$
\begin{aligned}
E[\log \frac{1}{p(x)}] \leq \log  E[\frac{1}{p(x)}]  = \log n
\end{aligned}
$$
where $n$ is the number of all possible events. Maximal value is attained when all possible events are equiprobable.

%%%%%%%%%%%%%%%%%%%%%%%%%%%%%%%%%%%%%%%%%%%%%%%%%%%%%%%%%%%%%%%%%%%%%%%%%%%%%%%
%%%%%%%%%%%%%%%%%%%%%%%%%%%%%%%%%%%%%%%%%%%%%%%%%%%%%%%%%%%%%%%%%%%%%%%%%%%%%%%

\end{document}